\documentclass[twoside,11pt]{article}
\usepackage{jair, theapa, rawfonts}

\ShortHeadings{Formally Verified Neurosymbolic Trajectory Learning via Tensor-based \ltlf{}}
{Chevallier, Smola, Schmoetten, \& Fleuriot}
\firstpageno{1}

\usepackage{amsfonts}
\usepackage{amsmath}
\usepackage{amssymb}
\usepackage{listings}
\usepackage[normalem]{ulem}
\usepackage{mathtools}

\usepackage[normalem]{ulem}
\usepackage{color,soul}
\usepackage{soul}
\usepackage{url}
\usepackage[utf8]{inputenc}
\usepackage{caption}
\usepackage{graphicx}
\usepackage{amsmath}
\usepackage{amsthm}
\usepackage{booktabs}
\usepackage{algorithmic}
\usepackage{subcaption}
\theoremstyle{plain}

\newtheorem*{theorem*}{Theorem}

\newcommand{\ltlf}{LTL$_f$}
\newcommand{\loss}{\ensuremath{\mathcal{L}}}
\newcommand{\dloss}{\ensuremath{d\loss}}

\usepackage{enumitem}
\usepackage{regexpatch}

\title{Formally Verified Neurosymbolic Trajectory Learning via Tensor-based Linear Temporal Logic on Finite Traces}

\author{\name Mark Chevallier \email mchevall@ed.ac.uk \\
       \name Filip Smola \email f.smola@ed.ac.uk \\
       \name Richard Schmoetten \email richard.schmoetten@ed.ac.uk \\
       \name Jacques D. Fleuriot \email jdf@ed.ac.uk}

\begin{document}

\maketitle

\begin{abstract}
We present a novel formalisation of tensor semantics for linear temporal logic on finite traces (\ltlf), with formal proofs of correctness carried out in the theorem prover Isabelle/HOL.
We demonstrate that this formalisation can be integrated into a neurosymbolic learning process by defining and verifying a differentiable loss function for the {\ltlf} constraints, and automatically generating an implementation that integrates with PyTorch.
We show that, by using this loss, the process learns to satisfy pre-specified logical constraints.
Our approach offers a fully rigorous framework for constrained training, eliminating many of the inherent risks of ad-hoc, manual implementations of logical aspects directly in an ``unsafe" programming language such as Python, while retaining efficiency in implementation. 
\end{abstract}

\section{Introduction}
\label{sIntro}

Neurosymbolic learning is an active and growing field of interest in artificial intelligence \cite{garcez2023neurosymbolic} that aims to combine the benefits of deep learning and symbolic reasoning. Of particular interest is the possibility to assist neural learning using symbolically encoded domain knowledge. This can be a complicated task and is prone to human errors in the translation process. In this work, we eliminate this risk by formally verifying the logic that implements the symbolic reasoning and automatically generating code that we then integrate into a neural network. We thus bring a high level of mathematical certainty to our implementation.

We use the proof assistant Isabelle/HOL \cite{nipkow2002isabelle} to formally specify and verify both the syntax and semantics of our logical language. Our approach is over tensors, the objects used for calculations in most deep learning frameworks. We also use Isabelle to generate executable code directly from our formalisation, guaranteeing that the implementation retains the formally verified correctness properties of the specification.

More concretely, we build a pipeline that integrates linear temporal logic over finite traces (\ltlf{}) \cite{de2013linear} formalised in Isabelle and deep learning in PyTorch \cite{paszke2019pytorch} via a novel interpretation of the \ltlf{} semantics over tensors. This allows us to automatically synthesise code for a loss function suitable to the application of any \ltlf{} logical constraint to the learning environment. While our solution is general and can be applied to any domain over which discrete temporal reasoning is appropriate, in this paper we demonstrate our pipeline with a set of motion-planning experiments. In these, we train a neural network to plan complicated paths fulfilling a variety of criteria expressible in \ltlf{}.  For an overview of our  Isabelle/HOL to PyTorch pipeline, see Figure \ref{fig:concepts} of Section \ref{ssPipeline}.

\subsection{Organisation of paper and contributions}

In the next section we discuss the background to our work, and briefly introduce \ltlf{}, tensors and the Isabelle/HOL theorem prover. Then, in Section \ref{sFormalisation} we present  a formal specification of \ltlf{} and some of its expected properties. We go on to specify a loss function \loss{} and its derivative \dloss{} that work over tensors representing finite traces of real-valued states varying in discrete time. We give an oveview of our formal proof that \loss{} is sound with respect to the semantics of \ltlf{}, and that \dloss{} formalises its derivative with respect to any variation in the input trace. We also describe our setup for the automatic generation of code for \loss{} and \dloss{} from their specifications using Isabelle's rigorous code extraction mechanism \cite{haftmann_codegen}. In Section \ref{sExperiments}, we illustrate an application of our general approach by considering a set of motion planning experiments.~Lastly, in Section \ref{sConclusion}, we discuss the significance of our work and potential future developments.

\medskip

\noindent The main contributions of this paper can be summarised as follows:
\begin{itemize}
    \item Formalised semantics of \ltlf{} over tensors.
    \item A formally-specified, differentiable loss function \loss{} over tensors that is sound with respect to the \ltlf{}  semantics.
    \item A formally verified, tensor-based derivative \dloss{} of the loss function.
    \item Automatically generated, reusable (OCaml) code for constrained, neural learning processes using \ltlf{}.
    \item Trajectory planning experiments demonstrating the practical applicability of our work.
    \item An empirical demonstration that the \ltlf{} constraint specification is domain-dependent, using a double loop path finding experiment.
\end{itemize}

\noindent Next, we briefly review the general background to our work before focusing on some of its more specific technical foundations.

\section{Background}
\label{sBack}

The \emph{standard} semantics for a logical syntax is a way of evaluating statements written in the language. This evaluation typically is either \emph{true} or \emph{false}, and takes place in a context where any variables used in the statements have values assigned to them. In our work, a context is the trace over which our logic is evaluated (discussed further in Section \ref{ssFormLtlf}). In what follows, let $[\![ \phi ]\!]_{L,i}$ denote the evaluation of a formula $\phi$ of a logical language $L$ under the standard semantics for $L$ in some context $i$.

A \emph{smooth} semantics over the same language $L$ can be considered a way of evaluating statements, also in some context, such that the evaluation is differentiable. Typically, the evaluation will give a real number that relates to the standard semantics. For some smooth semantics, e.g.~in fuzzy logic, this value may represent a confidence in the truth of the statement. For others, there may be a set of values corresponding to truth and others to falsehood. Note that smooth here means that the evaluation function is sufficiently differentiable for the purposes it is applied to -- in the context of learning processes, this means being once differentiable. We use $[\![ \phi ]\!]_{L,i}^*$ to denote the evaluation of formula $\phi$ of a logical language $L$ under a smooth semantics for $L$ in some context $i$.

\subsection{Learning using smooth semantics}
\label{ssBackRelated}

Fischer et al.~\citeyear{fischer2019dl2} have shown that it is possible to use a differentiable propositional logic to train a neural network . They use a simple logic to demonstrate that a loss function which is sound with respect to its semantics allows one to train a neural network to learn logical constraints. Our formalisation of atomic constraints is closely related to their work, although our treatment of other logical operators is different (see Section \ref{ssFormLtlf}).

Innes and Ramamoorthy \citeyear{innes2020elaborating_rss} extend Fischer et al.'s approach to use LTL and demonstrate that a path-planning neural network is able to learn to satisfy a subset of LTL constraints. However, the treatment of LTL in their paper does not make clear that they are working over finite traces in their task, which requires the different semantics of \ltlf{}. In particular, the treatment of \ltlf{} differs from LTL when it comes to the termination of a trace, as discussed in Section~\ref{ssBackLtlf} -- for example, the semantics for the Next operator is different from that in LTL. This means that the semantics they discuss in the paper cannot fully work over finite traces. Partly for this reason, the code supporting their work is incomplete and differs from the mathematical treatment discussed in the paper. Our work addresses these limitations (see Section~\ref{ssFormLtlf}).

This issue about the actual version of LTL being used, and the variance between the mathematical specification and its implementation motivated some of our earlier work on formally verifying the use of a \ltlf{}-based loss function to train neural networks~\cite{ChevallierWF22}. However, while rigorous, this previous work was not based on tensors, leading to inefficiencies in the generated code and the approach not scaling beyond simple examples. The current paper fully overhauls and extends this earlier effort. 

On the theorem proving side, Bentkamp et al.~\citeyear{bentkamp2019formal} formalised tensors in Isabelle/HOL as part of their work on proving the fundamental theorem of network capacity. As their formalisation was solely concerned with the mathematical proof of this theorem, it is impractical for generating efficient code. However, it did motivate our own approach to representing and formally reasoning about tensors (see Section~\ref{ssFormTensor}).

Other researchers have examined the semantics of different logics over tensors. How a subset of signal temporal logic (STL), a continuous time analogue to linear temporal logic, might be evaluated over tensor structures was examined by Leung et al.~\citeyear{leung2023backpropagation}. Their work describes the semantics with an algorithmic method rather than a simple loss function such as we detail in Section~\ref{sssFormLtlfFunc}, and is not formally verified. They were able to demonstrate their approach experimentally.

A smooth semantics for a variant of LTL (called Differentiable Temporal Logic or DTL) was adopted to learn task planning techniques by Xu et al.~\citeyear{xu2022don}. DTL is defined over a subset of a variant of LTL, and is differentiable and applicable to neural learning. The constants True and False are interpreted as positive and negative infinity, which may not be useful in a learning process. Similarly to the work by Innes and Ramamoorthy, it is unclear if DTL is intended to be defined over finite or infinite traces: the definition takes no account of trace termination behaviour. Moreover, unlike our own work, there are no formal guarantees of the properties of the smooth semantics.

The integration of a differentiable fuzzy logic into  neural network learning has also been proposed \cite{serafini2016learning}, with an implementation in TensorFlow \cite{BADREDDINE2022103649}. In contrast to our approach, which bridges the gap between formal theory and application through code generation, their implementation is completely separate from the theoretical description.

\subsection{Properties of smooth semantics}
\label{sssPropDiff}

Given the above, there are several important properties that we might want smooth semantics to possess:

\begin{enumerate}
    \item Compositionality: the smooth semantics should preserve structure of the standard semantics. Usually this is taken to mean that conjunction in the smooth semantics should be associative, commutative and idempotent.
    \item Soundness: $[\![ \phi ]\!]_{L,i}^*$ should output a value corresponding to truth if and only if $[\![ \phi ]\!]_{L,i}$ evaluates to true.
    \item Shadow-lifting~\cite{varnai2020robustness}: The value of a conjunction should reflect improvements in either of its conjuncts. In particular, if all conjuncts have comparable value, then an improvement in any of them should induce a similar improvement in the value of the conjunction. More formally, all partial derivatives should be positive:
    \[
    \forall n.\; \frac{\partial [\![ (x_1 \land x_2)]\!]_{L,i}^*}{\partial [\![ x_n ]\!]_{L,i}^*}\biggr|_{[\![ x_1 ]\!]_{L,i}^* = [\![ x_2 ]\!]_{L,i}^* = c} \ > 0 
    \]
    This means one can learn from partial improvements, even if not all conjuncts are improving simultaneously.
    \item Monotonicity: for all constraints $\phi_1$ and $\phi_2$, if $\phi_1$ entails $\phi_2$, then, if both are false under the standard semantics, $[\![ \phi_1 ]\!]_{L,i}^*$ should be no closer to the nearest value representing true than $[\![ \phi_2 ]\!]_{L,i}^*$. So, if the smooth semantics returns a value representing a loss, the weaker constraint produces a smaller loss than the stronger one\footnote{If both are true, then $[\![ \phi_2 ]\!]_{L,i}^*$ should be no closer to the nearest value representing false than $[\![ \phi_1 ]\!]_{L,i}^*$.}.
\end{enumerate}

\smallskip

In the literature, a uniform syntax has been proposed for such smooth semantics to potentially make it easier to mathematically verify the above properties~\cite{slusarz2022differentiable,slusarz2023logic} and it has been partly formalised in Coq~\cite{affeldt2024tamingdifferentiablelogicscoq}. The authors compared various existing smooth formalisms (e.g.\ DL2) against these properties and observed that these were obeyed to various degrees. In fact, no smooth semantics can possess all these desirable properties simultaneously, as the combination of idempotence and associativity (needed for compositionality) is incompatible with the shadow-lifting property as was observed by Varnai \& Dimarogonas \citeyear{varnai2020robustness}. We note that the work by {\'S}lusarz et al. does not cover a differentiable translation of LTL, finite or otherwise.

A set of properties has also been considered when dealing with a loss function for probabilistic classification models \cite{xu2018semantic}. Some of the properties they consider, e.g.\ soundness and monotonicity, are analogous to those discussed above. However, in this work, the logic is tailored to working for outputs that are probability distributions, so many of the properties are understandable only in this context.

In addition to some of the properties discussed above, the suitability of the derivative function to find minima has also been examined across several smooth semantics by Flinkow, Pearlmutter \& Monahan \citeyear{flinkow2024comparing}. The aim of this work was to assess whether these semantics produce good learning outcomes via gradient descent. They concluded that the magnitude of the derivatives produced (the steepness of the error curve) was more important to the learning process than other properties, such as shadow-lifting.

With regard to our work, we discuss our smooth semantics and formal proofs of soundness and compositionality in Section~\ref{sssFormLtlfFunc}. We also discuss shadow-lifting and monotonicity in Section~\ref{ssFuture}. In addition to these mathematical properties, a further important consideration when dealing with smooth semantics is how efficiently they can be implemented.
For example, we use tensors to improve the performance of code generated from our formalisation, but this needs to be accounted for in the formalisation (hence our standard and smooth tensor-based semantics for \ltlf{} as described in Section~\ref{sssFormLtlfEval}).
Additionally, we discuss how a minimal \ltlf{} language would lead to a less efficient loss function in Section~\ref{ssBackLtlf}. This emphasises that the aforementioned properties are only part of the picture of how a practical differentiable logic should be built.
    
\subsection{Linear temporal logic over finite traces}
\label{ssBackLtlf}

Linear temporal logic (LTL) allows the specification of constraints that can be evaluated over traces of states in discrete time \cite{pnueli1977}. In this context, a state is a set of measurements in a particular domain, and a trace is an ordered sequence of states, encoding how these change as time progresses.

Linear temporal logic constraints are either satisfied (true) or unsatisfied (false) at a specific moment within such a trace. Whether a constraint is satisfied may depend on how the trace changes in the future. \ltlf{} is a version of LTL where constraints are evaluated over traces that terminate at some finite point \cite{de2013linear}. {}In particular, this influences how the temporal operators work at the end of a trace.{} We assume that any constraint evaluated over an empty trace cannot be satisfied and must be interpreted as false.

\begin{figure*}[t!]
    \centering
    \includegraphics[width=0.5\textwidth]{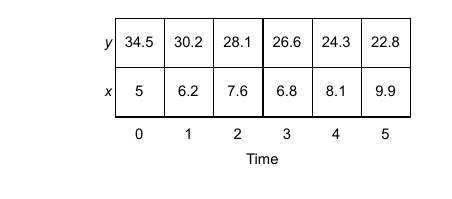}
    \caption{An example trace capturing the $x$ and $y$ coordinates of an agent.}
    \label{fig:trace}
\end{figure*}

The trace that an \ltlf{} constraint is evaluated over can be thought of as a finite matrix, as illustrated in Figure~\ref{fig:trace}. Consider columns as corresponding to a progression through time, and rows as corresponding to values that may change over time as the trace progresses. In our work, the elements of this matrix are always real numbers representing measurements of interest, such as the $x$ or $y$ coordinate of a time parameterised trajectory at a given moment. Indexing into the matrix extracts the value of a single measurement at a single point in time. 

From a more formal standpoint, constraints are syntactically represented by \ltlf{} formulae. The smallest constraint possible is an atomic one which extracts relevant measurements $\phi_n$ from the trace as real numbers and performs a comparison using them. For example, if we are monitoring the location of a robot in a two dimensional maze, we might be able to extract its $x$ and $y$ coordinates, and compare them against those for some goal. Inspired by Fischer et al.~\citeyear{fischer2019dl2}, we use the following comparisons as our atomic constraints: 

\begin{align*}
\sigma \Coloneq \phi_1\leq\phi_2 \mid \phi_1<\phi_2 \mid \phi_1=\phi_2 \mid \phi_1\neq\phi_2
\end{align*}

Building on these, the \ltlf{} formulae used to capture our more general constraints are:
\begin{align*}
\rho \Coloneq \sigma \mid \rho_1 \wedge \rho_2 \mid \rho_1 \vee \rho_2 \mid \mathcal{N}\rho \mid \mathcal{X}\rho \mid \square \rho \mid \Diamond \rho \mid \rho_1 \mathbin{\mathcal{U}} \rho_2 \mid \rho_1 \mathbin{\mathcal{R}} \rho_2 
\end{align*}

\noindent The first three are understood as per the syntax and semantics of propositional logic, with $\sigma$ being an atomic constraint (proposition), and $\rho_1 \mathbin{\wedge} \rho_2$ and $\rho_1 \mathbin{\vee} \rho_2$ representing conjunction and disjunction, respectively.

The remaining formulae introduce various temporal operators and can be (informally) understood as follows:
\begin{itemize}
	\item $\mathcal{N}\rho$ (Strong Next) holds if there is a next step and $\rho$ holds at that step.
    \item $\mathcal{X}\rho$ (Weak Next) holds if $\rho$ holds at the next step or there is no next step at all.
	\item $\square\rho$ (Always) holds if $\rho$ holds at the current step and all subsequent steps along the trace.
	\item $\Diamond\rho$ (Eventually) holds if $\rho$ holds at the current step or at least one subsequent step along the trace.
	\item $
 \rho_1\mathbin{\mathcal{U}}\rho_2$ (Weak Until) holds if $\rho_1$ holds at the current step and all the steps before $\rho_2$ starts holding, or holds at all subsequent steps if $\rho_2$ never holds.
	\item $\rho_1\mathbin{\mathcal{R}}\rho_2$ (Strong Release) holds if $\rho_1$ holds at some (possibly future) step and $\rho_2$ holds from the current step up to and including the point when $\rho_1$ starts holding.
\end{itemize}

We do not treat logical negation as a primitive operator in our work. Instead we specify a function $\textsc{Not}(\rho)$ that returns a constraint which we prove equivalent to the logical negation of $\rho$ (discussed in Section \ref{sssFormLtlfEval}). If we were to include logical negation as a primitive operator, it would mean we could not specify the loss function \loss{} using primitive recursion. This would prevent the use of inductive proofs over constraints, as we will discuss further in Section~\ref{ssFormLtlf}.

Given the above, our treatment of \ltlf{} is not minimal -- we could define the language with a smaller number of operators if we included logical negation. However, aside from the formalisation benefits, our approach also enables the generation of more efficient code: we do not need to define operators like Strong Release using a complicated chain of primitive operators that would slow down calculations during code execution. 

The above considerations demonstrate that a logically-informed but computationally aware approach to the specification of the symbolic language is important. 

\subsection{On Tensors}
\label{ssBackTensors}

Although tensors are fundamental to modern deep learning systems, their high-level representation and operations are usually taken for granted. In our case, though, these need to be made explicit to enable formal reasoning about their basic properties and more advanced notions based on them e.g.\ our loss function \loss{}. For this reason, we next give a brief overview of some of the aspects that are relevant to our current work, and delve a bit more deeply into their mathematical formalisation in Section \ref{ssFormTensor}.

Tensors are essentially multi-dimensional arrays of data that can be understood as a generalisation of vectors and matrices to arbitrary dimensions. In deep learning, tensors can be used to represent the input, intermediate computations, the weights of a neural network, and any output \cite{lim2013tensors}. Like a matrix, we can index into a tensor to obtain individual elements. Unlike a matrix, which has two indices, a tensor may take between zero and arbitrarily many indices. Following the terminology of Kolda and Bader \citeyear{kolda2009tensor}, the number of indices corresponds to the order of the tensor while the number of different values each index can take is the dimension (of that index).

\begin{figure*}[t!]
    \centering
    \includegraphics[width=0.7\textwidth]{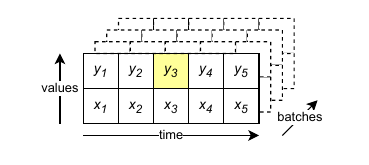}
    \caption{An example $5\times2\times4$ tensor representing a batch of \ltlf{} traces, highlighting the element indexed by (2,1,0).}
    \label{fig:tensor}
\end{figure*}

In our work, tensors are used to represent a trace over which we might evaluate some constraint in \ltlf{}. For example,  an order 3 tensor $T$ could be used to represent a batch of four traces of the $x$ and $y$ coordinates of a robot over five discrete time steps. So, if $T$ has size $5 \times 2\times4$, for instance, then the first dimension has a length of 5 representing the time; the second dimension has a length of 2 representing the $x$ and $y$ coordinates; and the third dimension indexes the traces that form the batch. The element indexed by $(2,1,0)$ (assuming indices start at $0$) would be the $y$ coordinate at the third time step from the first trace in the batch. In Figure~\ref{fig:tensor}, this element is highlighted in yellow.

\subsection{Isabelle/HOL}
\label{ssBackIsabelle}

For our work, the proof assistant Isabelle \cite{nipkow2002isabelle} provides the symbolic framework for formal specification, verification, and code generation. It is widely used for the formalisation of mathematics, especially real analysis and associated domains \cite{fleuriot2000,holzl2011three,immler2012numerical}, the verification of programs \cite{strecker2002formal,bulwahn2012smart,maric2010formal}, hardware \cite{tverdyshev2005combination}, and even software architecture \cite{marmsoler2018framework}.

Theories in Isabelle are collections of formal mathematical definitions (about e.g.\ types, algebraic objects, functions, or other objects), and theorems about their mathematical properties. Theorems and their proofs can be written in the pen-and-paper like, structured proof language Isar \cite{wenzel1999isar}. We use higher order logic \cite{gordon1988hol} in Isabelle, Isabelle/HOL, to formalise our work.

All theorems proven in Isabelle/HOL can be guaranteed to be valid. Informally, this is because Isabelle belongs to the LCF family of theorem provers \cite{milner1979lcf} which represent theorems as a type that must be built using a set of inference rules, starting from a small kernel of trusted axioms -- in our case, those of higher order logic. Moreover, Isabelle/HOL adheres to the so-called HOL methodology. This involves introducing new concepts via definitions rather than axiomatically; i.e.\ new theories are conservative extensions of existing ones \cite{pitts1993introduction,kunvcar2017safety}.  

Isabelle/HOL can generate code from formally specified functions, using a thin rewriting layer, into PolyML, OCaml, Haskell and Scala~\cite{haftmann_codegen}. This ability provides a link between Isabelle/HOL's formal concepts, such as our \loss{} and \dloss{} functions, and their code-generated versions. As discussed further in Section \ref{ssFormCodegen}, this provides various guarantees that the generated code respects the formally proven properties of its specification.

\section{Formalisation in Isabelle/HOL}
\label{sFormalisation}

We now discuss the details of our formalisation and proofs in three main sections: first we cover relevant parts of our tensor formalisation, then discuss our \ltlf{} work, and lastly, we briefly review code generation and how our formalisation feeds into it. Throughout, we use mathematical notation (rather than Isabelle syntax) to illustrate our specification and its properties.

\subsection{Tensors}
\label{ssFormTensor}

As previously mentioned, the use of tensors in a deep learning framework such as PyTorch is usually taken for granted. In our work, however, we need to give careful consideration to how these can be represented in formal mathematics and then used for code generation. 

Following Bentkamp et al.~\citeyear{bentkamp2019formal}, we represent an order $n$ tensor $T$ of dimensions ${d_0 \times d_1 \times \dots \times d_{n-1}}$ as a pair:
\begin{align*}
T=\langle(d_0,d_1,\dots,d_{n-1}), (e_0,e_1,\dots,e_{(\Pi_{i=0}^{n-1} d_i)-1})\rangle
\end{align*}

\noindent where $(d_0,d_1,\dots,d_{n-1})$ is an $n$-tuple of natural numbers representing the size of each dimension of $T$ and the ${(\Pi_{i=0}^{n-1} d_i)}$-tuple holds the elements of the tensor, which can be of any type $\mathbb{S}$. We define, Tensor$_\mathbb{S}$, the set of all tensors with elements of type $\mathbb{S}$:

\begin{align*}
\text{Tensor}_\mathbb{S}=\left\{\langle \vec{d},\vec{e}\rangle\: \middle|\: \exists n\in\mathbb{N}.\:\vec{d}\in\mathbb{N}^n \wedge \vec{e}\in \mathbb{S}^{(\Pi_{i=0}^{n-1} d_i)}\right\}
\end{align*}
\noindent where we use the notation $\vec{m}$ to represent a tuple. 

Note that we are defining a family of sets of tensors, parameterised by $\mathbb{S}$. For example, if $\vec{e}$ represents a tuple of rational numbers, we are defining $\text{Tensor}_{\mathbb{Q}}$. In Isabelle/HOL, there is no restriction on what $\mathbb{S}$ might be, but some theorems on tensors are only valid over certain element types. For instance, tensor addition is only defined for a type $\mathbb{S}$ that itself supports addition. In our work, we will only use tensors of boolean values and real numbers, which we denote by $\text{Tensor}_{\mathbb{B}}$ and $\text{Tensor}_{\mathbb{R}}$ respectively.

Each element $e_i$ is allocated to an indexed position within the tensor, described in terms of the tensor's dimensions. Element $e_0$ is allocated to $(0,0,\dots,0)$ in the tensor, and element $e_1$ is allocated to index $(0,0,\dots,1)$ (assuming the first dimension $d_0$ has a size of at least two).  $e_{d_0}$, the element whose index is equal to the size of the first dimension, is allocated to index $(0,\dots,1,0)$

For any $m\in\mathbb{N}$ where $\:m<(\Pi_{i=0}^{n-1} d_i)$, $e_m$ is allocated to a dimensional index as follows:

\begin{align*}
\left(m\text{ div } (\Pi_{i=1}^{n-1} d_i), ((m\mathbin{\text{mod}}(\Pi_{i=1}^{n-1} d_i)) \text{ div } (\Pi_{i=2}^{n-1} d_i),\dots\right)
\end{align*}

\noindent where $\text{div}$ is integer division.

In this way, we can translate element indices from a single natural number between $0$ and $(\Pi_{i=1}^{n-1} d_i)-1$, to indices using the dimensions of the tensor between $(0,0,\dots,0)$ and $(d_0-1,d_1-1,\dots,d_{n-1}-1)$, and back again. We use the function $\textsc{flatten}(\vec{i},\vec{d})$ to transition from a dimensional index to a linear one, and $\textsc{unflatten}(i,\vec{d})$ for the reverse. This is important as sometimes we need to treat the elements of a tensor as linearly indexed, and at other time as indexed via the tensor's dimensions. For example, $\textsc{flatten}((1,1)(2,3))=3$, and $\textsc{unflatten}(5,(2,3))=(1,2)$.

In subsequent sections, if we refer to an element $e_i$ where $i$ is a natural number, we are indexing it linearly, and if we refer to an element $e_{\vec{v}}$ where $\vec{v}$ is a tuple of natural numbers, we are indexing it via the tensor's dimensions.

\smallskip

We now introduce some of the basic operations used in our formalisation:

\begin{enumerate}
    \item $\textsc{dims}(\langle\vec{d},\vec{v}\rangle)$ returns $\vec{d}$, the tuple of dimension sizes.
    \item $\textsc{vec}(\langle\vec{d},\vec{v}\rangle)$ returns $\vec{v}$, the tuple of elements.
    \item $\vec{i}\triangleleft \vec{j}$ is an elementwise comparison between two tuples of natural numbers $\vec{i}$ and $\vec{j}$. It is true if and only if every element of $\vec{i}$ is less than the element in the corresponding position of $\vec{j}$, and both tuples are of equal length. Its intended purpose is to check if an index $\vec{i}$ being passed to a tensor $T$ refers to a valid element within that tensor, using $\vec{i}\triangleleft \textsc{dims}(T)$.
    \item $\textsc{lookup}(\vec{i},T)$ is a function that returns the element $e_{\vec{i}}$ from tensor $T$, assuming that $\vec{i}\triangleleft \textsc{dims}(T)$.
\end{enumerate}

To these, we add further tensor-based operations, namely \textsc{unop}, \textsc{binop}, \textsc{subt}, \textsc{backsubt}, and \textsc{replicate}. These are carefully formalised to allow for possible degenerate parameters since Isabelle/HOL functions must be total \cite{krauss2008defining} i.e.\ they must be defined for all their inputs. So, for example, we must be able to handle empty tensors properly.

\begin{enumerate}
    \item $\textsc{unop}(f,T)$ applies a unary function $f$ to every element of $T$: 
        \begin{align*}
            \textsc{unop}&(f,\langle\vec{d}, (e_0,\dots,e_n)\rangle)\\&=\langle\vec{d}, (f(e_0),\dots,f(e_n))\rangle
        \end{align*}
    \item $\textsc{binop}(f,T,G)$ applies the binary function $f$ to every pair of elements in matching positions in $T$ and $G$. For the result to be defined, $T$ and $G$ must be of equal size:
        \begin{align*}
            \textsc{binop} & (f,\langle\vec{d}, (e_0,\dots,e_n)\rangle, \langle\vec{d}, (e'_0,\dots,e'_n)\rangle)\\ & =\langle\vec{d}, (f(e_0,e'_0),\dots,f(e_n,e'_n))\rangle
        \end{align*}
    \noindent where $f$ denotes a function typically written between its operands, we will use the notation $T \circ_f G$ to represent $\textsc{binop}(f,T,G)$.
    \item $\textsc{subt}(\vec{i},T)$ takes an $m$-tuple $\vec{i}$ of natural numbers where $m$ is at most the order of $T$. It builds a subtensor of $T$ by fixing the first $m$ dimensional indices to those in $\vec{i}$. We define it by recursion on $\vec{i}$:
    \begin{align*}
        \textsc{subt}(\vec{i},T) = 
        \begin{cases}
            T & \text{ if }\vec{i}=()\\
            \textsc{subt} (i_1,\dots,i_m) \\
            \;\;\langle(d_1,\dots,d_{n-1}), \\
            \;\;\;\;(e_{i_0(\Pi_{j=1}^{n-1}d_j)+1},\dots,e_{i_0(\Pi_{j=1}^{n-1}d_j)+(\Pi_{j=1}^{n-1}d_j)})\rangle & \text{ if }\vec{i}=(i_0,i_1,\dots,i_m)
        \end{cases}
    \end{align*}
    \noindent where $T=\langle(d_0,d_1,\dots,d_{n-1}),(e_0,\dots,e_{(\Pi_{j=0}^{n-1}d_j)-1})\rangle$

    For example:
    \begin{align*}
        \textsc{subt}((2,17), \langle(6,22,3),(e_{(0,0,0)},\dots,e_{(5,21,2)})\rangle) = \langle(3),(e_{(2,17,0)},e_{(2,17,1)},e_{(2,17,2)})\rangle
    \end{align*}
    \item $\textsc{backsubt}(\vec{i},T)$ serves as the derivative of $\textsc{subt}(\vec{i},T)$, in the sense that it is an indicator function with respect to the elements of the tensor over which we are differentiating. It takes an $m$-tuple $\vec{i}$ of natural numbers where $m$ is at most the order of $T$. It returns a tensor of the same type and size, which has $1$ at every index that would be selected in the corresponding subtensor and $0$ everywhere else. This function is needed so that we can return the derivatives of functions which depend on $e_{\vec{i}}$ in the tensor. 
    \begin{align*}
        \textsc{backsubt} &((j_0,...,j_n), \langle(d_0,...,d_n,...,d_{n+m}),\vec{e}\rangle)\\
        & = \langle(d_0,...,d_n,...,d_{n+m}),\vec{e'}\rangle, \text{ where }e'_{(i_0,...,i_n,...,i_{n+m})}=\begin{cases}1 \text{, if }\forall k\leq n,i_k=j_k\\ 0\text{, otherwise}\end{cases}
    \end{align*}
    \noindent An illustrative example follows:
    \begin{align*}
        \textsc{backsubt} &((2,17), \langle(6,22,3),(e_{(0,0,0)},\dots,e_{(5,21,2)})\rangle)\\
        & = \langle(6,22,3),(e'_{(0,0,0)},\dots,e'_{(5,21,2)})\rangle, \text{ where }e'_{(j,k,l)}=\begin{cases}1 \text{, if }(j,k)=(2,17)\\ 0\text{, otherwise}\end{cases}
    \end{align*}
    \item $\textsc{replicate} (\vec{d},k)$ returns a tensor $T$ where $\textsc{dims}(T)=\vec{d}$ and every element of the tensor has the value $k$.
\end{enumerate}

\noindent With these operations established, we next discuss our tensor-based characterisation of \ltlf{}.

\subsection{\ltlf{} syntax and semantic formalisation}
\label{ssFormLtlf}

Here, we discuss the formalisation of \ltlf{} in Isabelle/HOL. We formalise the syntax of \ltlf{} following our definitions in Section \ref{ssBackLtlf}. In the rest of this section we will detail our novel formalisation of the semantics of \ltlf{} over tensors, both the truth value of an \ltlf{} constraint and its smooth semantics.

\subsubsection{Tensor-based semantics for \ltlf{}}
\label{sssFormLtlfEval}

A constraint has meaning under \ltlf{}: it is either true or false, given a time period to be evaluated over. In order for our formalisation to capture this, we define a function \textsc{eval} that can evaluate a \ltlf{} formula over a trace -- this effectively formalises the standard semantics of \ltlf{} over tensors.

\textsc{eval} takes three parameters: an \ltlf{ }constraint $\rho$, an order $n$ real tensor $T$ (with $n\geq2$), and a time parameter $t$, indicating the time step within each trace that $\rho$ should be evaluated from. $T$ contains the traces to be evaluated over, with dimensions beyond the first two representing the batching, and is indexable using $(i_0,i_1,\dots,i_{n-1})$. \textsc{eval} returns a boolean tensor (a member of $\text{Tensor}_\mathbb{B}$) of order $(n-2)$, where the element at dimensional index $(i_2, i_3, \dots , i_{n-1})$ is the truth value of $\rho$ when evaluated over the trace found at that element of the batch. Using the notation established in Section \ref{sssPropDiff}, $\textsc{eval}(\rho, T, 0)$ gives a tensor of results corresponding to $[\![ \rho ]\!]_{\text{\ltlf{}}, T}$.

In the rest of this section we examine \textsc{eval} by its cases, starting with the base case. If $t$ exceeds the maximum temporal index of the traces within $T$ then \textsc{eval} always evaluates to False, as we are evaluating $\rho$ over an empty trace. This is the first case, and overrides any other case. Note the use of the \textsc{replicate} function to ensure that the result is a tensor of the correct size.
\begin{align*}
    \textsc{ eval}(\rho,T,t) = \text{if }t\geq d_0,\text{ then }\textsc{replicate}((d_2,\dots,d_n),\text{False})
\end{align*}

If $t < d_0$, evaluation proceeds by induction along the constraint. An atomic constraint is evaluated depending on the precise comparison being used. Here, we use the \textsc{binop} function to perform the comparisons on an element-wise basis. The tensors we compare use the \textsc{subt} function to extract the precise values of interest (the value of each trace at time $t$ and measuring $\phi_1$ and $\phi_2$ within the state at that time).
\begin{align*}
    \textsc{eval}(\phi_1 < \phi_2,T,t) &= \textsc{subt}((t,\phi_1),T)\mathbin{\circ_<}\textsc{subt}((t,\phi_2),T))\\
    \textsc{eval}(\phi_1 \leq \phi_2,T,t) &= \textsc{subt}((t,\phi_1),T)\mathbin{\circ_\leq}\textsc{subt}((t,\phi_2),T))\\
    \textsc{eval}(\phi_1 = \phi_2,T,t) &= \textsc{subt}((t,\phi_1),T)\mathbin{\circ_=}\textsc{subt}((t,\phi_2),T))\\
    \textsc{eval}(\phi_1 \neq \phi_2,T,t) &= \textsc{subt}((t,\phi_1),T)\mathbin{\circ_{\neq}}\textsc{subt}((t,\phi_2),T))
\end{align*}

\noindent The other non-temporal constraints are evaluated straightforwardly:
\begin{align*}
    \textsc{eval}(\rho_1 \wedge \rho_2,T,t) &= \textsc{eval}(\rho_1,T,t)\mathbin{\circ_{\wedge}}\textsc{eval}(\rho_2,T,t)\\
    \textsc{eval}(\rho_1 \vee \rho_2,T,t) &= \textsc{eval}(\rho_1,T,t)\mathbin{\circ_{\vee}}\textsc{eval}(\rho_2,T,t)
\end{align*}

Temporal constraints are slightly more complicated to evaluate, depending on their behaviour upon reaching the end of the trace. In general, \textsc{eval} iterates down the trace for temporal constraints, by looking at $\textsc{eval}(\rho,T,t+1)$. However, if that would result in $t\geq d_0$, we need to stop that iteration and use either the base case discussed above or return True, depending on the constraint under consideration. Thus, for the following tensor-based \ltlf{} operators, we have:
\begin{align*}
    \textsc{eval}(\mathcal{N}\rho,T,t) = &\textsc{ eval}(\rho,T,t+1)\\
    \textsc{eval}(\mathcal{X}\rho,T,t) = &\text{ if }t=d_0-1\text{ then }\textsc{replicate}((d_2,\dots,d_n),\text{True})\\&\text{ else \textsc{eval}}(\rho,T,t+1)\\
    \textsc{eval}(\square\rho,T,t) = &\textsc{ eval}(\rho,T,t)\ \mathbin{\circ_{\wedge}} \\&\text{(if }t=d_0-1\text{ then }\textsc{replicate}((d_2,\dots,d_n),\text{True})\\&\text{ else \textsc{eval}}(\square\rho,T,t+1))\\
    \textsc{eval}(\Diamond\rho,T,t) = &\textsc{ eval}(\rho,T,t)\mathbin{\circ_{\vee}}\textsc{eval}(\Diamond\rho,T,t+1)\\
    \textsc{eval}(\rho_1\mathbin{\mathcal{U}}\rho_2,T,t) = &\textsc{ eval}(\rho_2,T,t)\mathbin{\circ_{\vee}}(\textsc{eval}(\rho_1,T,t)\mathbin{\circ_{\wedge}}\\
    &(\text{if }t=d_0-1\text{ then }\textsc{replicate}((d_2,\dots,d_n),\text{True})\\&\text{ else \textsc{eval}}(\rho_1\mathbin{\mathcal{U}}\rho_2,T,t+1)))\\
    \textsc{eval}(\rho_1\mathbin{\mathcal{R}}\rho_2,T,t) = &\text{ }(\textsc{eval}(\rho_1,T,t)\mathbin{\circ_{\wedge}}\textsc{eval}(\rho_2,T,t))\mathbin{\circ_{\vee}}\\
    &(\textsc{eval}(\rho_2,T,t)\mathbin{\circ_{\wedge}}\textsc{eval}(\rho_1\mathbin{\mathcal{R}}\rho_2,T,t+1))
\end{align*}

To illustrate how we handle the end of the trace, we discuss \textsc{eval}'s behaviour for constraints $\mathcal{N}\rho$ (Strong Next) and $\mathcal{X}\rho$ (Weak Next). As discussed in Section \ref{ssBackLtlf}, if these are evaluated on the very last time-step of a trace, Strong Next should return False and Weak Next should return True. When we evaluate at $t+1$, the correct behavior is captured for Strong Next by \textsc{eval}'s base case (the initial $t$-check), so the definition for $\mathcal{N}\rho$ is straightforward. To evaluate Weak Next, we check if $t=d_0-1$, in other words, we check if the evaluation is at the very last time-step of the trace. If so, we use the \textsc{replicate} function to return a True value for each trace in the batch, instead of iterating down the trace.

The Weak Until operator's formalisation is complicated, but worth examining to provide a detailed example of a temporal operator. $\rho_1 \mathbin{\mathcal{U}} \rho_2$ is True over a tensor $T$ at temporal position $t$ if and only if:

\begin{enumerate}
    \item $\rho_2$ is True over $T$ at $t$; \textit {or}
    \item \begin{enumerate}
        \item $\rho_1$ is True over $T$ at $t$; \textit{and}
        \item \begin{enumerate}
            \item $t$ is at the final time-step before the trace ends (checked using the method discussed previously); \textit{or} 
            \item ($\rho_1 \mathbin{\mathcal{U}} \rho_2$) is True over $T$ at temporal position $t+1$.
        \end{enumerate}
    \end{enumerate}
\end{enumerate}

\noindent This matches the behaviour we expect from the Weak Until operator as discussed in Section~\ref{ssBackLtlf}. If $\rho_2$ is not immediately True, then $\rho_1$ must be True and we can check the next time-step. If $\rho_2$ is never satisfied by the end of the trace, then $\rho_1 \mathbin{\mathcal{U}} \rho_2$ is True if and only if $\rho_1$ has been satisfied at every time step. If we reach a temporal position where $\rho_2$ is satisfied, $\rho_1 \mathbin{\mathcal{U}} \rho_2$ is True if and only if $\rho_1$ has been satisfied at every previous time step since we began. In the above, we focused on Weak Until as it is the most complicated case, but \textsc{eval} uses the same combination of checks, temporal iterations, and end-of-trace checking with all other temporal operators.

We formally prove basic theorems about the \textsc{eval} function, including:

\begin{align*}
    \textsc{eval}(\textsc{Not}(\rho),T,t) &= \textsc{unop}(\neg,\textsc{eval}(\rho,T,t)), \text{ if }t<d_0\\
    \textsc{eval}(\textsc{Not}(\textsc{Not}(\rho),T,t) &= \textsc{eval}(\rho,T,t)\\
    \textsc{eval}(\square (\square \rho), T,t) &= \textsc{eval}(\square \rho,T,t)\\
    \textsc{eval}(\mathcal{N} (\rho_1 \wedge \rho_2), T,t) &= \textsc{eval}(\mathcal{N} \rho_1 \wedge \mathcal{N} \rho_2,T,t)\\
    \textsc{eval}(\rho_1 \mathbin{\mathcal{U}} (\rho_1 \mathbin{\mathcal{U}} \rho_2), T,t) &= \textsc{eval}(\rho_1 \mathbin{\mathcal{U}} \rho_2, T,t)\\
    \textsc{eval}(\rho \wedge \mathcal{X} (\square \rho) ,T,t) &= \textsc{eval}(\square \rho, T,t)\\
    \textsc{eval}(\rho \vee \mathcal{N} (\Diamond \rho) ,T,t) &= \textsc{eval}(\Diamond \rho, T,t)
\end{align*}

\noindent These theorems establish that the \textsc{Not} function (as discussed in Section \ref{ssBackLtlf}) behaves properly and also demonstrate the expected \ltlf{} equivalences, thus showing that our tensor-based semantics matches the usual one.

\subsubsection{Smooth tensor-based semantics for \ltlf{}}
\label{sssFormLtlfFunc}

We next formalise the smooth semantics for \ltlf{} by defining the loss function \loss{} and its derivative \dloss{} over our tensors. \loss{} returns a real tensor (a member of $\text{Tensor}_\mathbb{R}$) of the same order as \textsc{eval}. 

\loss{} takes four parameters: the \ltlf{} constraint $\rho$, the tensor $T$, a temporal position $t$ we are to evaluate from, and a smoothing factor $\gamma$ used for differentiation (discussed in more detail below). Using the notation established in Section \ref{sssPropDiff}, $\loss(\rho, T, 0, \gamma)$ gives a tensor of results corresponding to $[\![ \rho ]\!]^*_{\text{\ltlf{}}, T}$. As discussed in that section, it is important that \loss{} is differentiable (with respect to the elements of the input tensor $T$) and that it is sound with respect to the standard semantics of \ltlf{} as formalised by the \textsc{eval} function. In this context, sound means that for a given constraint, trace, temporal position, and index in the result, \loss{} tends to zero as $\gamma$ does if and only if \textsc{eval} returns True given those same parameters. Simultaneously, the value for \loss{} is always non-negative in the limit $\gamma \to 0$, which is a property required for learning an objective through minimisation.

The differentiablity of \loss{} demands that the functions used to define it should be smooth\footnote{As noted in Section \ref{sssPropDiff}, by ``smooth'' here, we mean differentiable at least once, which is sufficient for our purposes.}. For example, we cannot use the $\max$ function in the definition of \loss{} as it is not differentiable everywhere.~As \loss{} defines a quantitative semantics for \ltlf{}, we need differentiable functions as analogues to the ones used to define the boolean, tensor-based \ltlf{} semantics given by \textsc{eval}.

All our smooth functions $f_\gamma$ of some underlying non-smooth $f$ have an additional real-valued parameter $\gamma$ representing the degree of smoothing. If $\gamma\leq0$, then  $f_\gamma = f$ i.e.\ the function is not smoothed. We thus define smooth binary versions of $\max$ and $\min$ \cite{cuturi2017soft}:

\begin{align*}
    \max\nolimits_\gamma(a,b) = \begin{cases}
        \max(a,b) &\text{ if }\gamma\leq0\\
        \gamma \ln(e^{a/\gamma}+e^{b/\gamma}) &\text{ if }\gamma>0\\
    \end{cases}
\end{align*}

\begin{align*}
    \min\nolimits_\gamma(a,b) = \begin{cases}
        \min(a,b) &\text{ if }\gamma\leq0\\
        -\gamma \ln(e^{-a/\gamma}+e^{-b/\gamma}) &\text{ if }\gamma>0\\
    \end{cases}
\end{align*}

\noindent and a Gaussian function that will be needed to define \loss{} for the atomic constraint $\phi_1\neq\phi_2$:

\begin{align*}
    \text{Gaussian}_\gamma(a) = \begin{cases}
        \text{if }a=0\text{ then }1\text{ else }0 &\text{ if }\gamma\leq0\\
        e^{-(a^2)/2\gamma^2} &\text{ if }\gamma>0\\
    \end{cases}
\end{align*}

In Isabelle, we formally prove that these functions are continuous,  each approaching its corresponding non-smooth variant as $\gamma$ tends to zero. When they are applied element-wise to tensors using the \textsc{unop} or \textsc{binop} functions, we denote them by \textsc{max}$_\gamma$, \textsc{min}$_\gamma$ and \textsc{gaussian}$_\gamma$, respectively.

Note that these functions are equal to their non-smooth variants when $\gamma \le 0$ and are differentiable only when $\gamma$ is positive. In practical terms, this means that there is a tension when setting $\gamma$: we wish it to be as low as possible, because some of the properties of the functions only hold when they are not smoothed,
but we cannot set $\gamma$ to zero, because then none of the smooth functions would be differentiable, and hence neither would \loss{}.
Moreover, implementation-wise, a complication can arise when operating with floating point numbers 
since values too close to zero can cause numerical instability (see Section~\ref{ssFormCodegen}). We discuss in Section \ref{sssContrained} how we arrive at useful values for $\gamma$ in an experiment context.

The function \loss{}, which returns a single real value per trace, is formally defined next, beginning with the base case:

\begin{align*}
    \mathcal{L}(\rho,T,t,\gamma) = &\text{ if }t\geq d_0 \text{ then }\textsc{replicate}((d_2,\dots,d_n),1)
\end{align*}

\noindent which is prioritised over all other conditions. This states that if we are evaluating at a time $t$ beyond the maximum temporal length of the trace $d_0$, then we return a value of 1, equivalent to falsehood. If the base case does not apply, the remaining cases recursively break down the constraint as follows:

\begin{align*}
    \mathcal{L}((\phi_1<\phi_2),T,t,\gamma) = &\textsc{max}_\gamma(\mathcal{L}((\phi_1\leq\phi_2),T,t,\gamma),\mathcal{L}((\phi_1\neq\phi_2),T,t,\gamma))\\
    \mathcal{L}((\phi_1\leq\phi_2),T,t,\gamma) = &\textsc{max}_\gamma((\textsc{subt}((t,\phi_1),T)\mathbin{\circ_{(-)}}\textsc{subt}((t,\phi_2),T)),\\&\textsc{ replicate}((d_2,\dots,d_n),0))\\
    \mathcal{L}((\phi_1=\phi_2),T,t,\gamma) = &\textsc{max}_\gamma(\mathcal{L}((\phi_1\leq\phi_2),T,t,\gamma),\mathcal{L}((\phi_2\leq\phi_1),T,t,\gamma))\\
    \mathcal{L}((\phi_1\neq\phi_2),T,t,\gamma) = &\textsc{gaussian}_\gamma(\textsc{subt}((t,\phi_1),T)\mathbin{\circ_{(-)}}\textsc{subt}((t,\phi_2),T)))\\
    \mathcal{L}(\rho_1\wedge\rho_2,T,t,\gamma) = &\textsc{max}_\gamma(\mathcal{L}(\rho_1,T,t,\gamma),\mathcal{L}(\rho_2,T,t,\gamma))\\
    \mathcal{L}(\rho_1\vee\rho_2,T,t,\gamma) = &\textsc{min}_\gamma(\mathcal{L}(\rho_1,T,t,\gamma),\mathcal{L}(\rho_2,T,t,\gamma))\\
    \mathcal{L}(\mathcal{N}\rho,T,t,\gamma) = &\text{ }\mathcal{L}(\rho,T,t+1,\gamma)\\
    \mathcal{L}(\mathcal{X}\rho,T,t,\gamma) = &\text{ if }t=d_0-1\text{ then \textsc{replicate}}((d_2,\dots,d_n),0)\\&\text{ else }\mathcal{L}(\rho,T,t+1,\gamma)\\
    \mathcal{L}(\square\rho,T,t,\gamma) = &\textsc{max}_\gamma(\mathcal{L}(\rho,T,t,\gamma),(\text{if }t=d_0-1\text{ then }\\&\textsc{ replicate}((d_2,\dots,d_n),0)\\&\text{ else }\mathcal{L}(\square\rho,T,t+1,\gamma)))\\
    \mathcal{L}(\Diamond\rho,T,t,\gamma) = &\textsc{min}_\gamma((\mathcal{L}(\rho,T,t,\gamma),\mathcal{L}(\Diamond\rho,T,t+1,\gamma))\\
    \mathcal{L}(\rho_1\mathbin{\mathcal{U}}\rho_2,T,t,\gamma) = &\textsc{min}_\gamma(\mathcal{L}(\rho_2,T,t,\gamma),\textsc{max}_\gamma(\mathcal{L}(\rho_1,T,t,\gamma),(\text{if }t=d_0-1\\&\text{ then \textsc{replicate}}((d_2,\dots,d_n),0)\\&\text{ else }\mathcal{L}(\rho_1\mathbin{\mathcal{U}}\rho_2,T,t+1,\gamma))))\\
    \mathcal{L}(\rho_1\mathbin{\mathcal{R}}\rho_2,T,t,\gamma) = &\textsc{min}_\gamma(\textsc{max}_\gamma(\mathcal{L}(\rho_1,T,t,\gamma),\mathcal{L}(\rho_2,T,t,\gamma)),\\&\textsc{max}_\gamma(\mathcal{L}(\rho_2,T,t,\gamma),\mathcal{L}(\rho_1\mathbin{\mathcal{R}}\rho_2,T,t+1,\gamma)))
\end{align*}

\noindent The structure of the \loss{} function closely follows that of the \textsc{eval} function. Throughout, where \textsc{eval} uses $\wedge$, \loss{} uses $\max_\gamma$. If both terms of this function evaluate to 0, corresponding to an \textsc{eval} value of True, then their maximum will also be 0, again corresponding to True. If either are positive, their maximum will also be positive, corresponding to False as we would expect. The argument for the correspondence between $\min_\gamma$ and $\vee$ proceeds similarly.

The atomic constraints deserve particular attention. We represent $\phi_1\leq\phi_2$ using ${\max_\gamma(\phi_1-\phi_2,0)}$. If $\phi_1 \leq \phi_2$, the difference will be zero or less, representing True, and if not, it will be positive representing False. We represent $\phi_1\neq\phi_2$ using the Gaussian function, which returns a positive value (representing False) if and only if the two values are equal, or nearly so. For the remaining constraints, we use $\max_\gamma$ combinations of the previous two: $\phi_1<\phi_2$ is $\phi_1\leq\phi_2$ and $\phi_1\neq\phi_2$, and $\phi_1 = \phi_2$ is $\phi_1\leq\phi_2$ and $\phi_2\leq\phi_1$.

Wherever the \textsc{eval} function uses \textsc{replicate} to return a True or False value, \loss{} uses \textsc{replicate} to return a 0 or 1, respectively. By translating from $\wedge$ and $\vee$ using $\max_\gamma$ and $\min_\gamma$, the atomic constraints detailed above, and this use of \textsc{replicate}, a simple correspondence can be drawn between how any of the \ltlf{} operators are treated under \textsc{eval} and \loss{}.

To illustrate this principle in detail, we will examine the Weak Until operator, following the example given in Section \ref{sssFormLtlfEval}. Recalling that $\max_\gamma(a,b) \to \max(a,b)$ and $\min_\gamma(a,b) \to \min(a,b)$ as $\gamma \to 0$, we can see that $\mathcal{L}(\rho_1 \mathbin{\mathcal{U}} \rho_2,T,t,\gamma)$ tends to zero as $\gamma$ does if and only if:

\begin{enumerate}
    \item $\mathcal{L}(\rho_2,T,t,\gamma)$ tends to zero as $\gamma$ does; \textit {or}
    \item \begin{enumerate}
        \item $\mathcal{L}(\rho_1,T,t,\gamma)$ tends to zero as $\gamma$ does; \textit{and}
        \item \begin{enumerate}
            \item $t$ is at the final time-step before the trace ends (\textsc{replicate} will return a value of 0 if this is true); \textit{or} 
            \item $\mathcal{L}(\rho_1 \mathbin{\mathcal{U}} \rho_2,T,t+1,\gamma)$ tends to zero as $\gamma$ does.
        \end{enumerate}
    \end{enumerate}
\end{enumerate}

A comparison of these conditions with those for \textsc{eval} precisely illustrates the simple correspondence between the two functions. This correspondence gives confidence that there is a clear relationship between the two that we can exploit to prove that \loss{} is sound with respect to \textsc{eval}. That is:

\begin{theorem*}[Soundness of \loss{}] \text{For every index $\vec{i}$ in the loss result}:

    \begin{enumerate}
        \item $\textsc{lookup}(\vec{i},\mathcal{L}(\rho,T,t,0))\geq 0$
        \item $(\textsc{lookup}(\vec{i},\mathcal{L}(\rho,T,t,\gamma))\to 0\text{ as }\gamma\to 0)\iff\textsc{lookup}(\vec{i},\textsc{eval}(\rho,T,t))$
    \end{enumerate}
    
\end{theorem*}
\noindent where \textsc{lookup} extracts the values (boolean for \textsc{eval} and real for \loss{}) returned in the result tensor.

The first part of the theorem states that the loss function \loss{} is always non-negative when $\gamma=0$, while the second part states that \loss{} tends to zero (for a given constraint, trace and time-step) as $\gamma$ does, if and only if the evaluation of that constraint over that trace and time-step is true. We formally establish this theorem in Isabelle/HOL, giving us guarantees about the correctness of our loss function \ltlf{}.~In particular,  the proof of the second property is split into demonstrating that \loss{} is continuous with respect to $\gamma$ and proving the following simpler property:

\begin{equation*}
    \textsc{lookup}(\vec{i},\mathcal{L}(\rho,T,t,0)) = 0 \iff \textsc{lookup}(\vec{i},\textsc{eval}(\rho,T,t))
\end{equation*}

\noindent which we prove by induction along the constraint $\rho$ and temporally down the trace $T$:

Beyond soundness, we also verify that the loss for conjunction and disjunction is idempotent, commutative and associative i.e.\ it is compositional (see Section \ref{sssPropDiff}). While commutativity and associativity hold no matter the value of the smoothing factor $\gamma$, idempotence holds only in the limit as $\gamma$ approaches zero. More formally, we have:

\begin{theorem*}[Idempotence, commutativity, associativity of \loss{}]~
    \begin{enumerate}[itemsep=0pt]
        \item For every index $\vec{i}$ in the loss result:\\
           $\lim_{\gamma\to 0} \left[\textsc{lookup}(\vec{i},\mathcal{L}(\rho \land \rho,T,t,\gamma)) - \textsc{lookup}(\vec{i},\mathcal{L}(\rho,T,t,\gamma))\right]=0$\\
           $\lim_{\gamma\to 0} \left[\textsc{lookup}(\vec{i},\mathcal{L}(\rho \lor \rho,T,t,\gamma)) - \textsc{lookup}(\vec{i},\mathcal{L}(\rho,T,t,\gamma))\right]=0$
        \item
            $\mathcal{L}(\rho_1 \land \rho_2,T,t,\gamma) = \mathcal{L}(\rho_2 \land \rho_1,T,t,\gamma)$\\
            $\mathcal{L}(\rho_1 \lor \rho_2,T,t,\gamma) = \mathcal{L}(\rho_2 \lor \rho_1,T,t,\gamma)$
        \item
            $\mathcal{L}(\rho_1 \land (\rho_2 \land \rho_3),T,t,\gamma) = \mathcal{L}((\rho_1 \land \rho_2) \land \rho_3,T,t,\gamma)$\\
            $\mathcal{L}(\rho_1 \lor (\rho_2 \lor \rho_3),T,t,\gamma) = \mathcal{L}((\rho_1 \lor \rho_2) \lor \rho_3,T,t,\gamma)$
    \end{enumerate}
\end{theorem*}
Next, we define \dloss{}. Our definition is based on the derivatives of our soft functions in combination with the chain rule. It takes the same parameters as \loss{}, but its output is a tensor of the same dimensions as the trace tensor $T$. The derivative of \loss{} with respect to any element $e_{\vec{i}}$ of $T$ is found at index $\vec{i}$ in the output of \dloss{}. We call this the derivative tensor of the function \loss{}.

The derivatives for the $\max_\gamma$ and $\min_\gamma$ functions use the chain rule, as noted above, because their parameters may themselves be the results of functions dependent on the input tensor. These derivative functions take four parameters: the two tensors that the underlying function is being used to compare and their two derivative tensors. For example, the function $\textsc{max}_\gamma(a, b)$ has a derivative calculated using the function $\textsc{dmax}_\gamma(a, da, b, db)$, where $da$ and $db$ are the derivative tensors for the functions used to calculate $a$ and $b$. In every case, we formally prove that this derivative is correct under the assumption that $da$ and $db$ are correct.

Derivatives for several of the atomic constraints use the \textsc{backsubt} function introduced in Section~\ref{ssFormTensor}. This produces the derivative tensor for \textsc{subt}, which is used in the \loss{} function for those constraints. 
For other atomic constraints, we build on this structure. For example, compare the definition of the derivative function for equality to the corresponding one given for the \loss{} function earlier:

\begin{align*}
    d\mathcal{L}((\phi_1=\phi_2),T,t,\gamma) = \textsc{dmax}_\gamma(&\mathcal{L}((\phi_1\leq\phi_2),T,t,\gamma),d\mathcal{L}((\phi_1\leq\phi_2),T,t,\gamma),\\&\mathcal{L}((\phi_2\leq\phi_1),T,t,\gamma),d\mathcal{L}((\phi_2\leq\phi_1),T,t,\gamma))
\end{align*}
\noindent Having proven that \textsc{dmax}$_\gamma$ is the derivative of $\textsc{max}_\gamma$ (discussed above), and given that we prove \dloss{} is defined correctly over the $\leq$ comparison, it is easily shown that $d\mathcal{L}((\phi_1=\phi_2),T,t,\gamma)$ is the derivative tensor for \loss{}$((\phi_1=\phi_2),T,t,\gamma)$.

As another example, we also examine the definition of \dloss{} with respect to the Weak Until operator:
\begin{align*}
    d\mathcal{L}(\rho_1 \mathbin{\mathcal{U}} \rho_2,T,t,\gamma) =\ &
    \textsc{dmin}_\gamma(\\
    &\:\:\mathcal{L}(\rho_2,T,t,\gamma),\\
    &\:\:d\mathcal{L}(\rho_2,T,t,\gamma),\\
    &\:\:\textsc{max}_\gamma(\\&\:\:\:\:\mathcal{L}(\rho_1,T,t,\gamma),\\
    &\:\:\:\:\text{ if }t=d_0-1\text{ then }\textsc{replicate}((d_2,...,d_n),0)\\&\:\:\:\:\:\:\text{ else }\mathcal{L}(\rho_1 \mathbin{\mathcal{U}} \rho_2,T,t+1,\gamma)),\\
    &\:\:\textsc{dmax}_\gamma(\\
    &\:\:\:\:\mathcal{L}(\rho_1,T,t,\gamma),\\
    &\:\:\:\:d\mathcal{L}(\rho_1,T,t,\gamma),\\
    &\:\:\:\:\text{if }t=d_0-1\text{ then }\textsc{replicate}((d_2,...,d_n),0)\\&\:\:\:\:\:\:\text{else }\mathcal{L}(\rho_1 \mathbin{\mathcal{U}} \rho_2,T,t+1,\gamma),\\
    &\:\:\:\:\text{if }t=d_0-1\text{ then }\textsc{replicate}((d_0,...,d_n),0)\\&\:\:\:\:\:\:\text{else }d\mathcal{L}(\rho_1 \mathbin{\mathcal{U}} \rho_2,T,t+1,\gamma)))
\end{align*}

As can be seen, this is a nontrivial definition.  We invite the reader to compare the above case to the corresponding one for \loss{}, to convince themselves that, given  \textsc{dmin}$_\gamma$ and \textsc{dmax}$_\gamma$ are derivatives of \textsc{min}$_\gamma$ and \textsc{max}$_\gamma$ respectively, this definition is the derivative of \loss{} for the Weak Until operator. We fully verify this property in Isabelle/HOL for all the operators.

The above definition is only one part of the entire \dloss{} definition, making it understandable why its manual implementation may be error-prone. Our approach eschews this weakness by automatically generating the code for the formally verified \dloss{}.

\subsection{Code generation in Isabelle/HOL}
\label{ssFormCodegen}

We obtain code from our formalisation using Isabelle/HOL's code generation facilities~\cite{haftmann_codegen}. 
This is a formal program synthesis method that passes our mathematical specifications through a thin translation layer to a target programming language, in our case OCaml. The translation layer of the code generator, analogous to the kernel of the theorem prover, uses a small number of trusted code translations to build types and function definitions.

As already mentioned, this ensures that the generated code retains the mathematical properties that we have proven for our formal specification. This principle is fundamental to our pipeline.

For several of our functions, we expand on the formal specifications using \textit{code equations}. Consider some function $\mathcal{F}$ defined in Isabelle/HOL, with a set of properties proven against it. The definition of $\mathcal{F}$ used for these proofs may be ideally suited for theorem proving purposes, but may be inefficient when translated using code generation. A code equation for $\mathcal{F}$ is then a formally proven equivalent, alternative definition for $\mathcal{F}$. This alternative definition is designed to generate more efficient code. As an example, consider \textsc{subt}, which is defined recursively in Section \ref{ssFormTensor}. While this definition makes proofs easier, it can result in many recursive calls if used for calculations over large tensors. An alternative definition of \textsc{subt} instead  directly selects a \textsc{section} of elements starting and ending at computed linear indices in the original element array:
\begin{align*}
    \textsc{subt}((i_0,i_1,...i_m),\langle(d_0,d_1,...,d_{m+n}),\vec{e}\rangle) = &\langle(d_{m+1},...,d_{m+n}),\textsc{section}(\vec{e})\rangle
\end{align*}    

\noindent This is formally proven to be equivalent to the original definition but eschews recursion and thus generates more efficient code.

Code equations can also prove useful for dealing with the limitations imposed on us by translating between the types of Isabelle/HOL and those used in the target language. For example, when dealing with integers in Isabelle/HOL, there is no need to consider any limits on the type. However, in code, the number of bytes allocated to an integer typed value restricts the values we can compute with.

This above issue is more complicated when dealing with real numbers as these are mapped to floating point numbers in code and may result in edge cases where a property proven to be true over the mathematical reals may not be generally true over floating point numbers. This is an issue with any program that uses floating point types and is not unique to our pipeline.

In general, the range of values for a given type should be sufficient for any application we are dealing with, but we may wish to redefine some functions that can generate very large or very small values, even if only in an intermediate stage of execution. This is more likely with our functions if the $\gamma$ hyperparameter is very low. We can deal with this via code equations as well. As an example, we use the following code equation for $\max_\gamma$ (compare with the original definition given in Section \ref{sssFormLtlfFunc}):
\begin{align*}
    \max_\gamma(a,b)=\begin{cases}
        \max(a,b) &\text{ if }\gamma\leq0 \\
        \gamma \cdot ({a}/{\gamma}+\ln(1+e^{(b-a)/\gamma}))&\text{ if } \gamma>0, a<b\\
        \gamma \cdot ({b}/{\gamma}+\ln(1+e^{(a-b)/\gamma}))&\text{ if } \gamma>0, b<a\\
        \gamma \cdot ({a}/{\gamma}+\ln 2)&\text{ if } \gamma>0, a=b\\
    \end{cases}
\end{align*}

This definition has more cases and is thus more difficult to work with when proving properties. However, it works with a greater range of parameters than the original definition, where, for instance, if $\gamma$ is very small, $e^{a/\gamma}$ can lead to an overflow during computation. The code equation version reduces the numerator of the fraction we exponentiate with. This means that there are fewer combinations of parameters that will lead to out of bounds errors when working with the code generated version that uses floating points. Thus we can have a wider range of usable parameters under this definition.

Using similar approaches, we generate code for our entire specification that both works efficiently and helps circumvent issues with numerical computations.

\subsection{Integration with PyTorch}
\label{ssPipeline}

\begin{figure}[ht]
    \centering
    \includegraphics[width=0.8\textwidth]{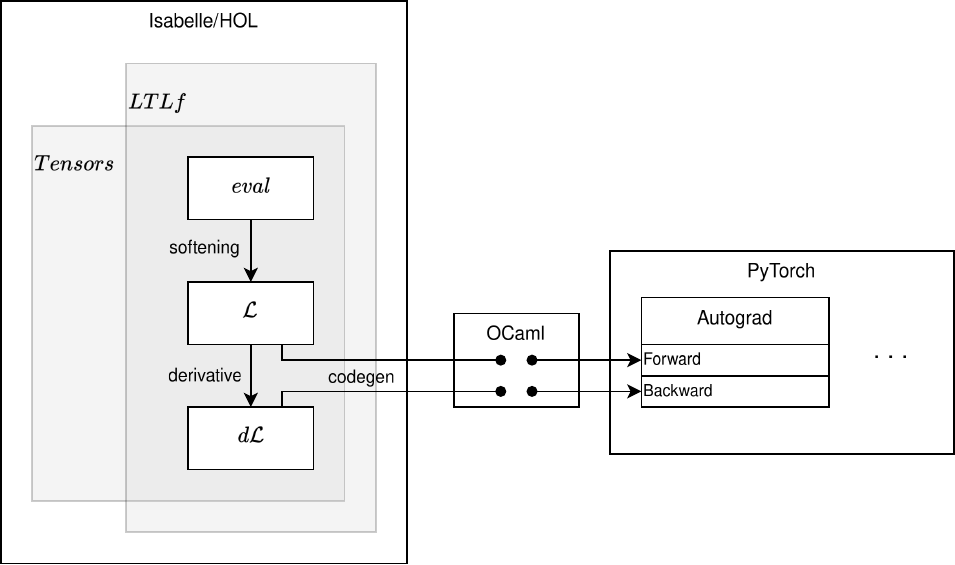}
    \caption{Isabelle-PyTorch neurosymbolic pipeline.}
    \label{fig:concepts}
\end{figure}

In our experiments, the generated OCaml code is integrated into PyTorch using its autograd engine. The \loss{} function is used in the forward pass of an optimiser, and the \dloss{} function in the backward pass. Figure \ref{fig:concepts} shows the overall pipeline, with the formalisation and proofs  established in Isabelle/HOL, the code generated in OCaml, which acts as the bridge into PyTorch used for our experiments.

The OCaml code is not PyTorch specific, i.e.\ it can be used in any other framework that accepts executable programs implementing such forward and backward functions. We chose PyTorch due to the ease with which our generated code  can be integrated and also because it is a widely used platform.

Moreover, we note that our OCaml code is not neural network specific either: it is usable with any algorithm that learns to minimise a function via gradient descent as we describe in the next section.

\section{Experiments}
\label{sExperiments}

To show that our pipeline produces usable learning outcomes, we give a demonstration of our \ltlf{} loss function being used to directly optimise trajectories in Section~\ref{ssExperDirect}, and then look at its use in a deep learning setting involving dynamic movement primitives in Section~\ref{ssExperDomain}.

\subsection{Direct optimisation of trajectories}
\label{ssExperDirect}

The setting of these introductory demonstrations is that of finding a two-dimensional trajectory that satisfies some constraint expressed in \ltlf{}. In our first example, a trajectory is represented as a vector in $(\mathbb{R}\times\mathbb{R})^{N}$, where $N$ is the number of discrete timesteps. This vector acts as the trace our \ltlf{} constraints are evaluated over. The initial trajectory is a straight line (with a small random component to avoid poor learning due to symmetry) from $(0,0)$ to $(1,1)$, with constant velocity, and $N=50$ (the time interval between timesteps is $dt=1$).
We optimise $N-2$ points on this trajectory, excepting the first and last timesteps (to fix the endpoints of the trajectory), using PyTorch.

The first constraint we add avoids a disk of radius $0.1$ around the point $o=(0.4,0.4)$. The \ltlf{} constraint for this task is: $\square (0.1\leq \|p - o\|)$, where $p$ is a position on the trajectory, and $\|\cdot\|$ is the Euclidean metric.

Real-world trajectories often need to obey additional constraints. For safety, the trajectory may need to obey a fixed speed limit, such as $s = 0.15$. Using dot notation for time derivatives, the velocity of the trajectory is $\dot p$, and we can express this speed constraint as $\square (s\geq \|\dot p\|)$, where our discretisation of the velocity $\dot p$ is
$\dot p_t = (p_{t}-p_{t-1})/{dt}$,
for $0<t\leq N$.

Additionally, any robot or vehicle following a trajectory may have some maximal acceleration, e.g.\ maximal thrust in the case of a rocket. Thus we restrict our example trajectory to have acceleration of at most $a = 0.02$, by constraining it with $\square (a\geq \|\ddot p\|)$, with $\ddot p_t = (\dot p_t - \dot p_{t-1}) / dt$ for $1 < t \leq N$.

\begin{figure}[t!]
    \centering
    \begin{subfigure}{0.48\textwidth}
    \includegraphics[width=\textwidth]{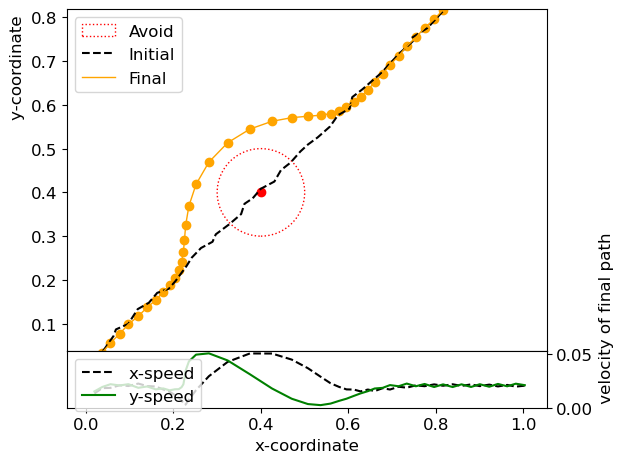}
    \caption{Trajectory and speed components for a simple trajectory constrained to avoid a circular region (outline dotted in red).}
    \label{fig:direct-avoid}
    \end{subfigure}
    \begin{subfigure}{0.48\textwidth}
    \includegraphics[width=\textwidth]{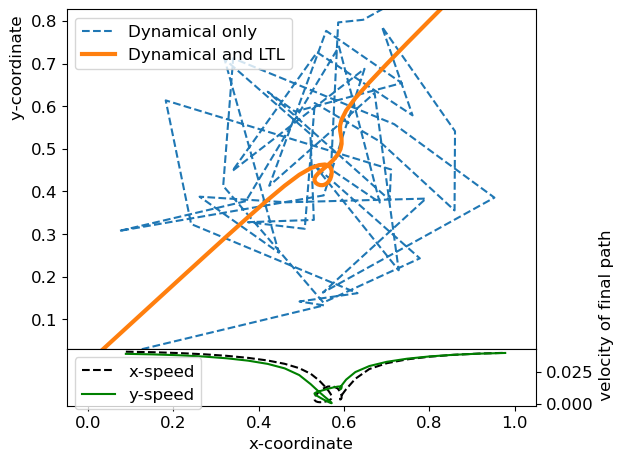}
    \caption{Joint optimisation of a random initial set of points and velocities. Augmenting the dynamical loss with a well-chosen \ltlf{} term leads to a smoother trajectory.}
    \label{fig:direct-smoothen}
    \end{subfigure}
    \caption{Trajectories and velocities for demonstrations with simple trajectories.
    }
    \label{fig:direct-1}
\end{figure}

In summary, we optimise a trajectory (represented in $(\mathbb{R}\times\mathbb{R})^{N}$) under the loss function generated from a simple constraint expressed purely in \ltlf{}: \[ \square \big( (0.1 \leq \|p - o\|) \wedge (s \geq \|\dot p\|) \wedge (a \geq \|\ddot p\|) \big) \] The result of this simple training procedure is shown in Figure~\ref{fig:direct-avoid}. Note how the speed and acceleration constraints lead to a shallow, wide arc around the region to avoid. Without such constraints, the trajectory simply speeds up enough to traverse the region around $(0.4,0.4)$ in a single timestep. If no samples inside this region are evaluated, the loss for avoiding point $o$ is 0, and thus the constraint is satisfied by the trace, even when this is not a satisfactory trajectory as envisaged originally.

But the loss function used for optimisation does not have to be purely formulated in \ltlf{}. We can, for example, add an \ltlf{} constraint to a pre-existing loss term, and optimise this mixed loss function to obtain more refined results. Indeed, consider the setting where a trajectory is represented not only by a list of coordinate samples, but also includes the speed along each axis at each sample.
Now our trace representing the trajectory is a long vector $(p, q)$ concatenated from coordinate samples $p = (x_t, y_t)_{0 \leq t \leq N}$ and speed samples $q = (v_t, u_t)_{0 < t \leq N}$, so that $(p,q) \in \mathbb{R}^{4N -2}$. In order to be a valid trajectory, our trace must obey \[\frac{x_t - x_{t-1}}{dt} = v_t\;\text{ and }\;\frac{y_t - y_{t-1}}{dt} = u_t\;\text{ for all }\;0 < t \leq N\] All of these dynamical conditions can be collected into the matrix form $E \times (p,q)^T = 0$, where each row of $E$ contains non-zero indices only where they extract the desired components of $(p,q)$.
As an example, the condition for $v_1$ would be \[\underbrace{(1, 0, -1, \dots, dt, \dots)}_{\text{first row of }E} \times {\underbrace{(x_0, y_0, x_1, \dots, v_1, \dots)}_{(p,q)}}^T = x_0 - x_1 + v_1 dt = 0 \]

\noindent We add four equations, i.e.\ four last rows to $E$, that specify the endpoints $(x_0, y_0) = (0,0)$ and $(x_N, y_N) = (1,1)$.

Minimising the function $D(p,q) = E \times (p,q)^T$ merely ensures the trace is representative of a well-formed trajectory, not necessarily a trajectory that is reasonable for any practical purpose.
Since it is rare we start with as good a guess of our trajectory as we did with the straight line, we start with a uniformly random trace, optimise the dynamical loss $D$ and show the result in Figure~\ref{fig:direct-smoothen} (dotted blue line). While this trajectory may contain matching coordinates and velocities, it remains somewhat erratic.

To address this, and obtain a smoother trajectory, we formulate some additional requirements using \ltlf{}, and perform optimisation (from the same random initial state) of a weighted sum of the loss function $D$ and the loss generated from the \ltlf{} constraint. The constraint we use (as an illustrative example only) specifies upper and lower bounds for the components of the velocity:
\[ \square \left(
  v \geq 0 \land
  u \geq 0 \land
  v \leq 0.03 \land
  u \leq 0.03
\right)
\]
This constraint is satisfied by a trajectory whose velocity at each point has positive components less than $0.03$; i.e.\ the velocity is bounded componentwise and points towards the upper right of Figure~\ref{fig:direct-smoothen}.

The result of this experiment is shown in Figure~\ref{fig:direct-smoothen}. The trajectory optimised with an \ltlf{} term varies less wildly, and is smoother, with smaller oscillations in velocity than the trace trained without the \ltlf{} constraint. The specific trajectories shown in Figure~\ref{fig:direct-1} should be taken only as representative examples.  Notice we obtain a smoother trajectory even though the optimisation only finds a local minimum, rather than fully satisfying the \ltlf{} constraint! 

The remainder of our experiments try to get away from this simple setting, and move further towards using \ltlf{} constraints jointly with more realistic training algorithms, specifically when learning from data. This new model is introduced in the next section, and constraints and experiments are explained in more detail thereafter.

\subsection{Neural optimisation of trajectories  using Dynamic Movement Primitives}
\label{ssExperDomain}

The environment within which each of the following experiments takes place is a plane. A feed-forward neural network learns the parameters for a Dynamic Movement Primitive (DMP) \cite{schaal2005learning} across this plane.

DMPs span a wide class of smooth trajectories that approach a goal state from arbitrary initial conditions. A trajectory $y(t)$ approaching a goal $y_{\mathrm{final}}$ is modelled as a damped spring with a forcing term $f$, can be summarised as the following equation adapted from Ijspeert et al.~\citeyear{ijspeert2013}:
\[
\ddot y = \alpha (y_{\mathrm{final}}-y) - \beta \dot y + f
\]
for constants $\alpha$ and $\beta$. The forcing term $f$ is a function of the initial conditions of the trajectory, learned from data as discussed in Innes \& Ramamoorthy \citeyear{innes2020elaborating_rss}.
Once the forcing term and parameters are fitted to data, a set of states along the trajectory can be produced by evolving the above equation from an initial state in discrete time.

DMPs are frequently used to represent complex movements, and present many advantages over the simpler model of Section~\ref{ssExperDirect}. They represent a trajectory abstractly: as such, the learned trajectory is used to generate a set of samples for any list of time stamps. For point-set trajectories, the model and the sample are equal, and the timestamps are considered fixed across the entire problem. More importantly, the learnable parameters of the DMP are variables in a dynamical system, rather than just an independent set of positions (and perhaps velocities). Thus gradient descent influences the dynamics of the entire trajectory, while for point-set trajectories any dynamical meaning must be added manually (e.g.\ through the dynamical constraint in Section~\ref{ssExperDirect}).

Once the initial, unconstrained, DMP has been learned we repeat the learning process, injecting our loss function \loss{} for an \ltlf{} constraint into the learning process, and observe how the resulting trajectory seeks to satisfy the constraint as we would expect. The constraint is evaluated against the trajectory produced by the DMP, not the DMP itself directly. However, as the trajectory is differentiable with respect to the DMP's parameters, the loss function can still be used to adjust the neural network's weights using backpropagation.

The training data follow a trajectory from which we extract $N=100$ sequenced points in the plane describing a curve with small random perturbations, simulating a demonstrator in a supervised learning process following a trajectory from the origin to a destination in the plane. The DMP learns this trajectory, which we subsequently alter using \ltlf{} constraints.

\subsubsection{Unconstrained training}
\label{sssUnconstrained}

Let $D$ be the trajectory of the demonstrator along the curve and let $Q$ be the trajectory produced by the neural network, with both trajectories expressed using paired $x$ and $y$ coordinates. $Q$ and $D$ are $N \times 2$ matrices. Let $Q_i$ be the row vector at index $i$ of $Q$ (similarly for $D_i$ and $D$). The per-sample imitation loss, $L_d$, for this sample pair is given by:

\begin{equation*}
    L_d(Q,D) = \frac{1}{N}\sum_{i=0}^{N-1}{\|Q_i - D_i\|^2}
\end{equation*}

Intuitively, $L_d$ penalises the learned trajectory for deviating from the demonstrator. By minimising this imitation loss, the neural network learns to produce a trajectory matching the demonstrator's. If we are using batches of samples, we calculate the average imitation loss for learning purposes. Initially, $L_d$ is used as the sole loss in the training of the neural network over 200 epochs using the Adam optimizer \cite{adam} with a learning rate of $10^{-3}$.

We carry out this unconstrained training to contrast the learned behaviour of the neural network without any constraint with the one involving the injection of our \ltlf{}-based loss.

\subsubsection{Constrained training with \ltlf{}}
\label{sssContrained}

Recall that $Q$, the trajectory of the DMP, encodes the $x$ and $y$ coordinates at each time-step. Let $P$ be the trace used to evaluate an \ltlf{} constraint, which may additionally encode derived values about the trajectory, such as velocity or distance from a certain point, as well as any constants used for comparison.

To use \loss{} in the learning process, we define a differentiable function $g$ that maps $Q$ to the trace $P$, an \ltlf{} constraint $\rho$ over that trace, and a smoothing factor $\gamma > 0$. We use the function $g$ to transform the $x$ and $y$ coordinates of $Q$ to the values we are constraining e.g.\ the distance from a particular point, as discussed in Section \ref{sssExperimentDescription}. We then use this constraint in the learning process by changing the per-sample loss function to: 

$$L_{full} = L_d(Q,D) + \eta\cdot\mathcal{L}(\rho,g(Q),\gamma)$$ 
where $\eta$ is a positive real number representing the weighting of the constraint violation loss against the imitation loss. This weighting is adjustable, allowing us to change the priority of satisfying the constraints relative to the imitation data. The loss function \loss{} is implemented by the generated code from Isabelle/HOL.

We now repeat the same training procedure as for the unconstrained case, but with this augmented loss function $L_{full}$. We use $\eta=1.0$, the default weighting. For $\gamma$, we proceeded empirically by  by testing our most complicated scenario, involving the Compound constraint (see below for the full description),  with different $\gamma$ values from 0.5 to 0.001. The lowest value to produce stable results was $\gamma=0.005$.

\subsubsection{Description of Experiments}
\label{sssExperimentDescription}

We lay out 5 different problems:
\begin{enumerate}
    \item \textbf{Avoid}: The trajectory (always) avoids a disk of radius $0.1$ around the point $o=(0.4,0.4)$. We compute the Euclidean distance between the trajectory at the current time, which we denote as $p$,  and $o$. The \ltlf{} constraint is: $\square (0.1\leq \|p-o\|)$.
    \item \textbf{Patrol}: The trajectory eventually reaches $o_1=(0.2,0.4)$ and $o_2=(0.85,0.6)$ in the plane. With the same notation as above, this constraint becomes: $(\Diamond (\|p-o_1\|~\leq~0)) \wedge (\Diamond(\|p-o_2\|~\leq~0))$.
    Note that we do not use the comparison $=$ as the Euclidean distances are non-negative and this formulation has a lower computational cost.
    \item \textbf{Until}: The $y$ co-ordinate, $p_y$, of the trajectory cannot exceed $0.4$ until its $x$ co-ordinate, $p_x$, is at least $0.6$: $(p_y\leq 0.4)\text{ }\mathcal{U}\text{ }(0.6\leq p_x)$.
    \item \textbf{Compound}: A more complicated test combining several conditions. The trajectory should avoid a disk of radius $0.1$ around the point $o_1=(0.5,0.5)$, while eventually touching the point $o_2=(0.7,0.5)$. Further, the $y$ co-ordinate of the trajectory should not exceed $0.8$. With the same notation as previous experiments, this compound constraint is represented as: $(\square (0.1\leq \|p-o_1\|)) \wedge(\Diamond (\|p-o_2\| \leq 0))\wedge(\square (p_y\leq0.8))$.
    \item \textbf{Loop}: The trajectory eventually reaches $o_2=(0.4,0.8)$ and then eventually reaches $o_1=(0.2,0.6)$. To do this, it must loop back on and intersect itself, forming a loop that represents a large divergence from its unconstrained behaviour. The \ltlf{} constraint is: $\Diamond (\|p-o_2\|\leq0 \wedge (\mathcal{X} (\Diamond (\|p-o_1\|\leq0))))$
\end{enumerate}

To further illustrate the purpose of the function $g$, consider the Avoid test. Here, $g$ is acts row-wise on the trajectory $Q$, producing new row vectors whose elements are the Euclidean distance $p_{do}$ and the constant $0.1$, as these are the only values reasoned over by the \ltlf{} constraint. For the other tests, $g$ likewise acts of $Q$ to produce the values reasoned over by each constraint.

\subsubsection{Results}
\label{sssExperResults}

\begin{table}[t!]
\centering
 \begin{tabular}{|c c c c c|} 
 \hline
 Test & $L_d$ & $\mathcal{L}$ & $L_{full}$ & \\ [0.5ex] 
 \hline\hline
 Unconstrained & 0.0587 & --- & 0.0587 & \\
 Avoid & 0.0681 & 0.0263 & 0.0944 & \\
 Patrol & 0.0694 & 0.0448 & 0.1142 & \\
 Until & 0.0688 & 0.0193 & 0.0881 & \\
 Compound & 0.0807 & 0.0284 & 0.1092 & \\ 
 Loop & 0.0800 & 0.0486 & 0.1286 & \\ [1ex]
 \hline
 \end{tabular}
 \caption{Imitation loss ($L_d$), constraint loss (\loss{}), and total loss ($L_{full}$) for each experiment.}
 \label{table:numresults}
\end{table}

We run each experiment for 500 epochs and record their loss results against a set of validation data (a demonstrator trajectory) as shown in Table~\ref{table:numresults}.

The losses show that when we perform constrained training, we understandably have higher imitation losses than we do for unconstrained training. This is because as we adjust the trajectory to satisfy the constraint, we are deviating further from the demonstrator trajectory. The two sources of loss, constraint and imitation, cannot both be reduced to zero.

\begin{figure*}[t!]
    \centering
    \includegraphics[width=0.3\textwidth]{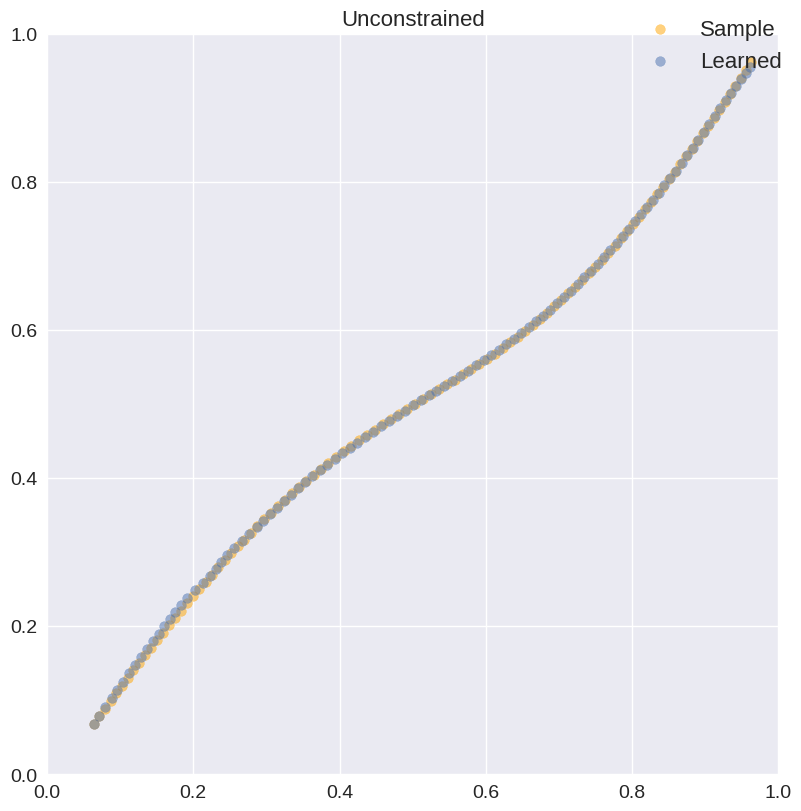}
    \includegraphics[width=0.3\textwidth]{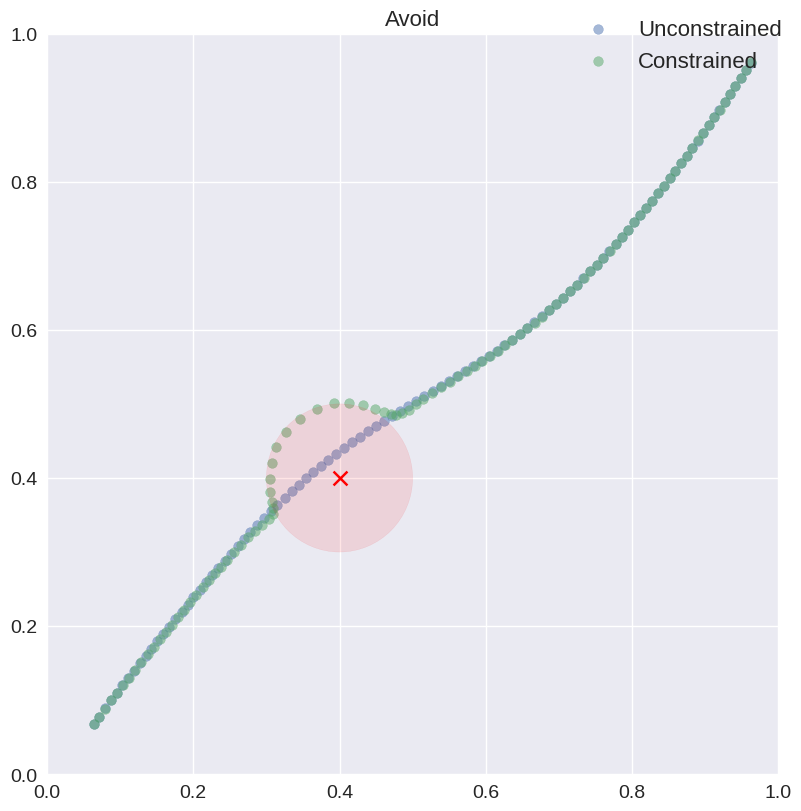}
    \includegraphics[width=0.3\textwidth]{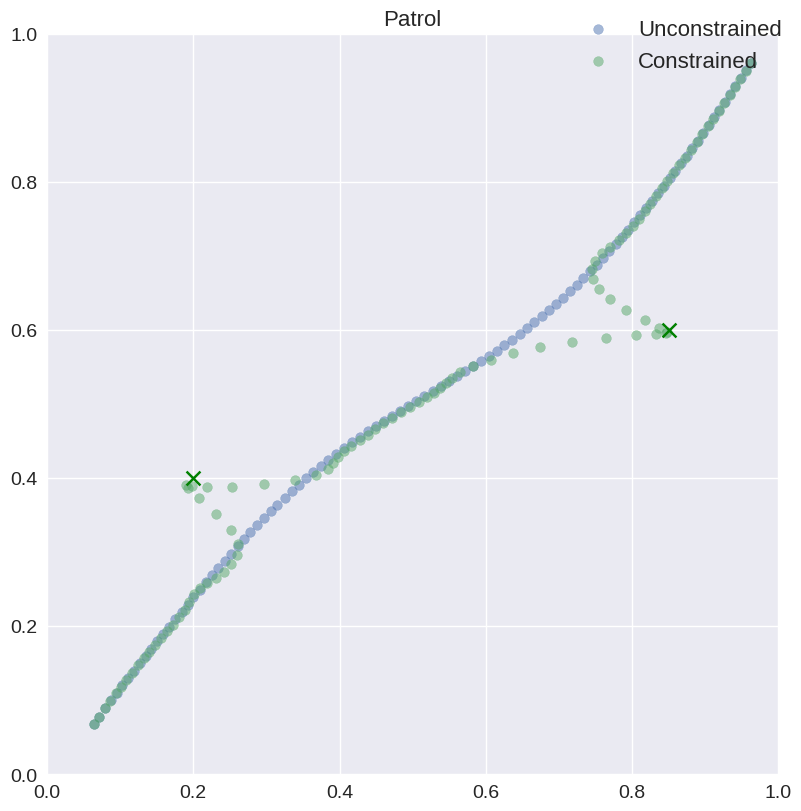}
    \includegraphics[width=0.3\textwidth]{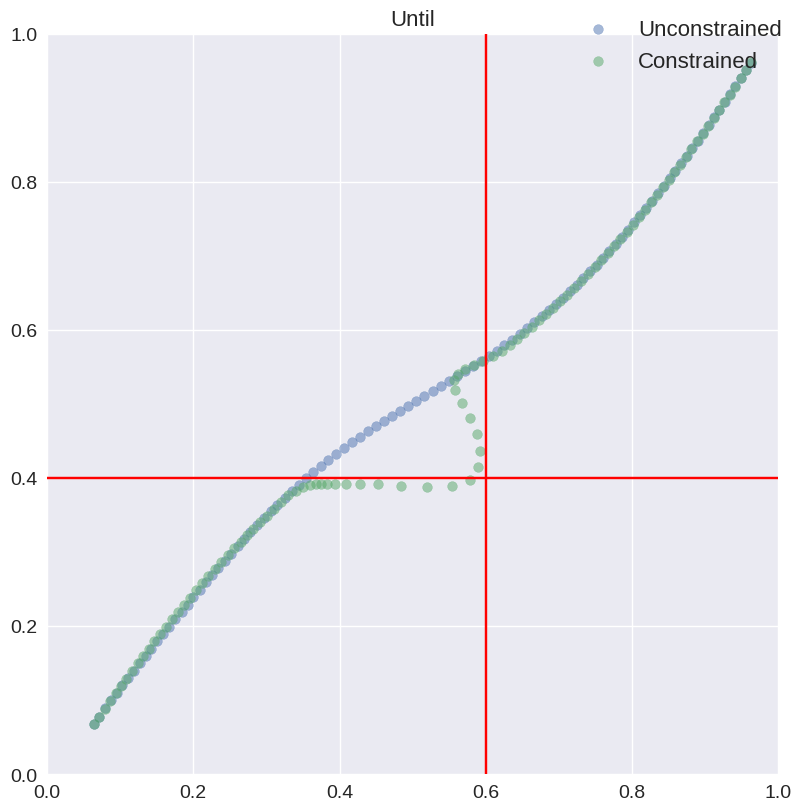}
    \includegraphics[width=0.3\textwidth]{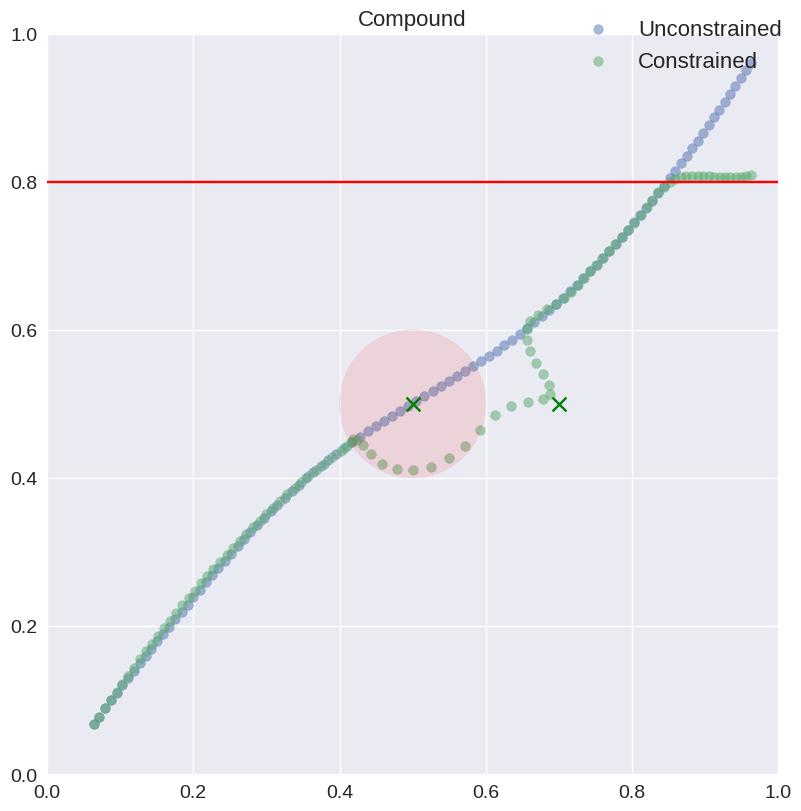}
    \includegraphics[width=0.3\textwidth]{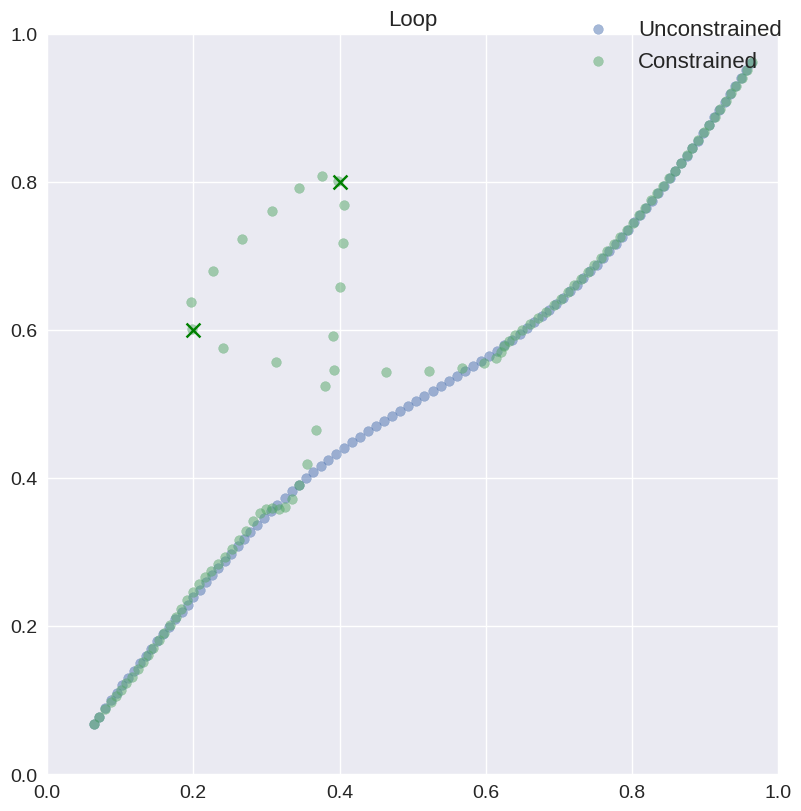}
    \caption{Trajectories for the Unconstrained, Avoid, Patrol, Until, Compound and Loop experiments.}
    \label{fig:results}
\end{figure*}

We should avoid direct comparisons of the losses from different constrained tests, as constraints may be measuring different aspects. The values are not really comparable beyond saying that any zero value represents complete satisfaction of a constraint. Even though in most of our experiments the constraints are satisfied (as is visually observable from the graphs of trajectories in Figure~\ref{fig:results}), the constraint losses are not zero. Given that we are dealing with a soft \loss{} function with a positive $\gamma$ parameter, this is expected. By the soundness property (described in Section~\ref{sssFormLtlfFunc}), we would expect the constraint loss of fully satisfied constraints to tend to zero as $\gamma$ does.

Figure~\ref{fig:results} shows the the unconstrained trajectories (in blue) and the trajectories modified by the constraint (in green), and, in each case, clearly demonstrates a new trajectory that tries to meet the associated constraint. Slight violations may happen because, in addition to the constraint loss, the trajectory loss plays a part in forming the neural network's training. These losses work against each other in our tests. It is also clear that the trajectories depicted are smoother than those generated in Section \ref{ssExperDirect}. Overall these results demonstrate the clear effectiveness of the logical constraint in changing the learned behaviour of the DMP, and the efficacy of using DMPs to describe trajectories. 

As a final remark here, we note the efficiency benefit of defining the semantics of \ltlf{} over tensors. An earlier version of this pipeline worked with losses evaluated directly against a trace with no tensor context, instead calculating over scalar values \cite{ChevallierWF22}. The changes made to that work to formalise a loss function against tensors produce speed improvement which are at least ten fold for the same experiments as in the previous work.

\subsection{Constraint relaxation}
\label{ssDoubleloop}

Calculating losses for the constraints describing complicated trajectories can become time consuming. This is especially true if we have multiple nested temporal operators. However, it is sometimes possible to simplify the logical constraint by choosing an appropriate representation of the problem. 

We demonstrate the above by extending the loop experiment previously described in Section \ref{sssExperimentDescription} to one that loops twice, once towards the top left of the demonstrated trajectory, and once to the bottom right. This requires an increase in the number of nested temporal operators to capture the new loop constraint. The obvious extension for the  constraint to produce two consecutive loops is to introduce two additional points. Ordered by their $x$ coordinate, we thus set four points: $o_1=(0.2,0.6)$, $o_2=(0.4,0.8)$, $o_3=(0.6,0.2)$ and $o_4=(0.8,0.4)$ to be visited in the order $o_2, o_1, o_4, o_3$.
The extended constraint then becomes:
\[\Diamond (\|p-o_2\|\leq0 \wedge \mathcal{X} (\Diamond (\|p-o_1\|\leq0) \wedge \mathcal{X} (\Diamond (\|p-o_4\|\leq0) \wedge \mathcal{X} (\Diamond (\|p-o_3\|\leq0))))).\]

Note that this constraint nests four Eventually operators within each other. This operator is evaluated by recursing through states of the trace until the condition is satisfied (or the trace is exhausted). Since Eventually is nested four times, the evaluation will check the operator with an upper bound of $t^4$ times, where $t$ is the number of time steps in the trace. This nesting of temporal operators is costly in terms of the number of comparisons it performs. 

We next argue that we can relax this constraint to one that is more tractable when in our DMP setting. Consider instead  the conjoined constraint:
\[\Diamond (\|p-o_2\|\leq0 \wedge \mathcal{X} (\Diamond (\|p-o_1\|\leq0))) \wedge \Diamond (\|p-o_4\|\leq0 \wedge \mathcal{X} (\Diamond (\|p-o_3\|\leq0))).\] 
This still involves four Eventually operators, but this time nested only twice, and then joined by a conjunction. In natural language, this says the path will reach points $o_2$, then $o_1$, in order. Separately, the path will also reach point $o_4$, and then $o_3$. It differs from the previous constraint as it can be satisfied if the path proceeds through the points in the following order: $o_2,\,o_4,\,o_3,\,o_1$, or in several other ways that do not describe the desired double loop.

\begin{figure}[t!]
    \centering
    \begin{subfigure}{0.48\textwidth}
    \includegraphics[width=\textwidth]{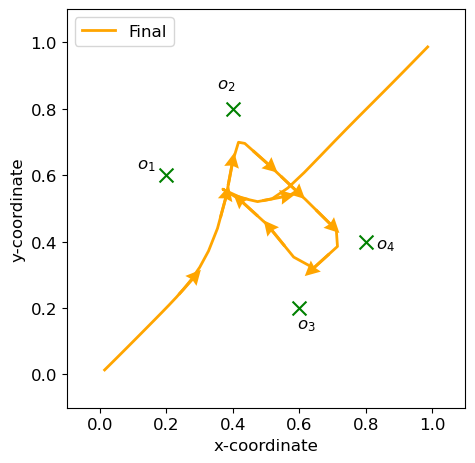}
    \caption{Conjoined constraint.}
    \label{fig:direct-double-conj}
    \end{subfigure}
    \begin{subfigure}{0.48\textwidth}
    \includegraphics[width=\textwidth]{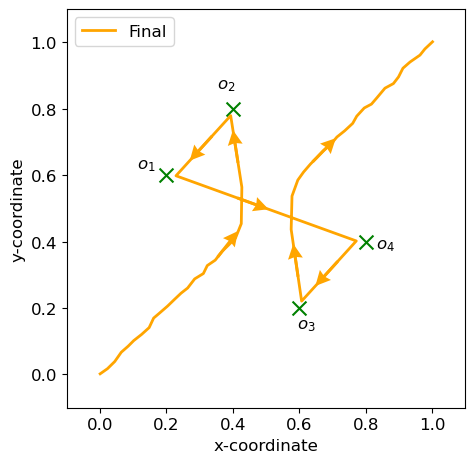}
    \caption{Nested constraint.}
    \label{fig:direct-double-nest}
    \end{subfigure}
    \caption{Example trajectories represented as simple sets of coordinates and speeds, trained using conjoined and nested constraints respectively to follow two loops. The conjoined constraint may not lead to desired behaviour for this representation (depending on initial state, random factors in optimisation procedures, etc.), in contrast to using DMPs. Observe the speed limit is effective only before reaching the first and after reaching the last points.}
    \label{fig:direct-double}
\end{figure}

This behaviour is in fact what we observe when training the simple model, where trajectories are sets of coordinates and speeds (see Section~\ref{ssExperDirect}). A comparison of nested and conjoined constraints in this setting is given in Figure~\ref{fig:direct-double}. It is notable that despite imposing a constraint encouraging slower speed, the point-based representation is unable to satisfy this speed limit across the entire trajectory.

We surmise this is because the constraint ``eventually, reach some point'' can be satisfied by a single point, and therefore only leads to gradients for individual points. As no other trajectory satisfying this constraint is reachable by following available gradients, the higher speed-limit loss is balanced by a low point-reaching loss in a local minimum. The nested constraint leads to significantly slower training, taking over 70 times as long per iteration, but leads to a double loop as desired (with the violation of speed constraints described above).

\begin{figure}[t!]
    \centering
    \includegraphics[width=0.5\textwidth]{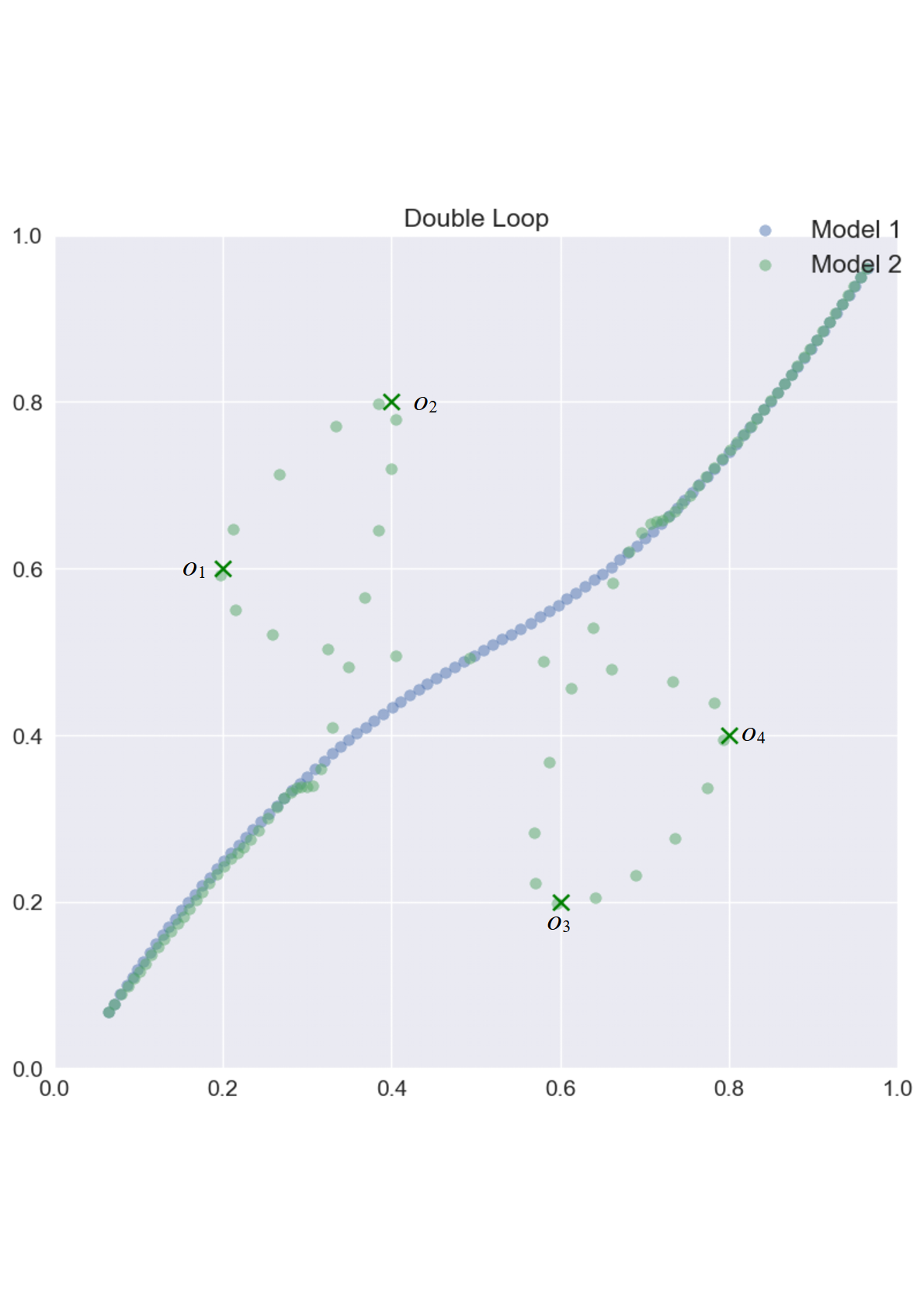}
    \caption{Conjoined double loop constraint using DMPs. The conjoined constraint, although capable of being satisfied in a number of ways, follows the double loop because we define the trajectory using a DMP.}
    \label{figdoubleloop}
\end{figure}

However, when used with the DMP, the double loop is consistently achieved with the conjoined constraint, as shown in Figure \ref{figdoubleloop}. There are two differences between the two sets of experiments: 
\begin{enumerate}
    \item The DMP experiments, in addition to learning to prevent a breach of the constraint, also learn to imitate a demonstrator (as discussed in Section \ref{ssExperDomain}).
    \item
    The trajectory for the simple model is defined directly by a list of coordinate pairs.
    For the DMP experiments, the trajectory is defined by the parameters of a set of differential equations produced by the neural network, a much more sophisticated method.
\end{enumerate}

We performed some ablation experiments with both methods, adding an imitation loss to the simple model, and removing it from the DMP,
and were able to conclude that it did not contribute to the successful double loop of the DMP experiment under the conjoined constraint. We conclude that the reason for this success is that the DMP is a more sophisticated means of representing a trajectory. This additional sophistication, likely by encouraging low curvature, works in combination with the logical constraint to produce the desired result.

The conjoined constraint, with its reduced nesting of Eventually operators, requires fewer calls to the evaluation function when it recurses down a trace. In fact, instead of $t^4$, the upper bound on state evaluations is $2t^2$, a significant reduction for almost all $t$. In practical terms, this means that the experiment concludes significantly faster than for the fully nested constraint.

Apart from the efficiency gains in this specific case, a positive lesson arising from this experiment is that the representation of the problem domain can have a bearing on the \ltlf{}-based loss function. In particular, a more appropriate representation of the trajectory as a DMP allows us to use a simpler, easily evaluated \ltlf{} constraint to achieve a desired behaviour, even if by itself it does not provide a complete specification of the intended behaviour.

\section{Conclusion}
\label{sConclusion}

We conclude this paper by giving an overview of current and potential further work, and provide some final reflections on our research contribution.

\subsection{Future work}
\label{ssFuture}

We have discussed a formal, tensor-based formulation of \ltlf{} that can be used for neurosymbolic learning. However, there are further logics that we might wish to work with, that allow for a broader range of constraints. For example, we are currently working on formalising signal temporal logic~\cite{donze2013signal} and integrating it into our neurosymbolic framework. More generally, our work  indicates that the approach could be successfully and practically applied to arbitrary logics, defining smooth semantics and proving them sound with respect to the logic's standard one.

It would be interesting to further study the loss landscape generated by our smooth semantics, through properties such as the shadow-lifting property (see Section~\ref{sssPropDiff}). Any proofs of properties of the smooth semantics, be it regarding exploration or soundness, directly describe the behaviour of optimisation problems. One could then explore ideas such as annealing the value of $\gamma$ throughout the learning process, reducing it as the learning process converges on the underlying function, improving soundness.

In the same context, one might investigate the property of monotonicity (see Section~\ref{sssPropDiff}), i.e.\ whether
logical entailment between constraints is compatible with the ordering of their loss values. This may require a different approach to the loss function, to ensure logical contradictions are always assigned greater loss than unsatisfied comparisons.

While in this paper we demonstrate our \ltlf{} loss function on path and motion planning, it is effectively applicable to any temporal domain, e.g.\ automated planning of schedules. Furthermore, the experimental domains are, at present, outside of the formal environment. In future work we aim to investigate the benefits of including the domain in the formalisation, with the aim of formalising and verifying domain-dependent loss function simplification.

Another potential avenue for future work would involve implementing the neural network in the same language into which we generate code from our formalisation. This would eliminate the need to transfer tensors and constraints to and from Python, enabling tighter integration of generated code with the neural network. OCaml, which we generate in our present work, has a machine learning library providing bindings for LibTorch~\cite{ocamltensors}. Isabelle can also generate code in Haskell, for which there exists a similar library called Hasktorch\footnote{Hosted at \texttt{http://hasktorch.org}}.

\subsection{Final Observations}
\label{finalobs}

We have presented a fully formal tensor-based formulation of \ltlf{} and discussed how this can be used to specify a loss function and its derivative, and to generate code that is integrated into PyTorch for logic-based optimisation and neural learning.

Our experimental results show that these constraints can successfully change the training process to match the desired behaviour. In contrast to ad-hoc, manual implementation in Python, our formally verified approach provides strong guarantees of correctness through a combination of proof and faithful code generation.

Throughout, the decisions we have taken have been informed by the requirement to accurately model  \ltlf{}, but additionally by the (sometimes competing) requirements of formal proof and efficient code generation. Formal proof is greatly assisted by a logical structure that allows for easy induction over both constraints and tensors.

We have also seen how the choice of domain representation can impact on the applicability of constraints to achieve a goal. In our application to path modelling, a sophisticated trajectory model based on DMPs allows us to simplify constraints in a way that a na\"ive coordinate-based approach does not.

Our research emphasises several important lessons:

\begin{enumerate}
    \item The soundness of the smooth semantics is 
    important, but it is not the only property that should be satisfied.
    \item Other properties such as compositionality and shadow-lifting (discussed in Section~\ref{ssBackRelated}) are important, but need not be perfectly met for the logic and its loss function to be useful for effective learning.
    \item It is important to pay attention to how functions are specified for ease of proof, and how those specifications can be reworked into verified, equivalent versions for the generation of efficient code.
\end{enumerate}

Overall, we believe that our work demonstrates an effective and trustworthy approach to using logical constraints in learning. It allows for the fully formal specification of a smooth semantics and proof of important mathematical properties, the automatic generation of faithful code for learning, and thus strong confidence that the learning agent will behave as specified. 

\subsection*{Acknowledgements}
This research is funded by the Edinburgh Laboratory for Integrated Artificial Intelligence (ELIAI) and the Fonds National de la Recherche, Luxembourg (AFR 15671644).

\bibliography{constrainedtraining}
\bibliographystyle{theapa}

\end{document}